\documentclass{article}
\usepackage[table]{xcolor}
\usepackage{hhline}
\usepackage{xspace} 
\newcommand{\algname}{Inner Monologue\xspace}
\usepackage{todonotes}
\usepackage{algorithm}
\usepackage{booktabs}
\usepackage{graphicx} 
\usepackage{algpseudocode}
\usepackage{siunitx}
\usepackage{svg}
\usepackage[moderate]{savetrees}  % [subtle,moderate,extreme]
\definecolor{lightgray}{gray}{0.9}
\usepackage{listings}
\usepackage[font=small,labelfont=bf,tableposition=top]{caption}
\usepackage{subcaption}

\usepackage[font=footnotesize]{caption}
\usepackage[compact]{titlesec}
\titlespacing{\section}{1pt}{*0}{*0}
\titlespacing{\subsection}{1pt}{*0}{*0}
\titlespacing{\subsubsection}{1pt}{*0}{*0}

\usepackage{listings}
\usepackage{xcolor}
\usepackage[normalem]{ulem}
\usepackage{soul}

\definecolor{Cerulean}{rgb}{0,0,0.95}
\definecolor{LimeGreen}{rgb}{0.15,0.65,0.15}
\definecolor{RoyalBlue}{rgb}{0.25,0.41,0.88}
\definecolor{success_detection_color}{rgb}{0.56, 0.427, 0.957}
\definecolor{scene_description_color}{rgb}{0.416, 0.659, 0.31}
\definecolor{vqa_color}{rgb}{0.067, 0.333, 0.80}
\definecolor{robot_action_color}{rgb}{1.0, 0.60, 0}

\definecolor{Rose}{rgb}{1.0, 0.15, 0.21}
\definecolor{Orange}{rgb}{1.0, 0.5, 0.0}
\definecolor{Gray}{gray}{0.6}
\definecolor{Black}{gray}{0.0}
\definecolor{Purple}{rgb}{0.77,0.12,0.64}
\definecolor{codegreen}{rgb}{0,0.8,0}
\definecolor{codered}{rgb}{0.95,0,0.3}
\definecolor{codegray}{rgb}{0.5,0.5,0.5}
\definecolor{codepurple}{rgb}{0.58,0,0.82}
\definecolor{backcolour}{rgb}{0.95,0.95,0.95}
 
\lstdefinestyle{Python}{
    language        = Python,
    basicstyle      = \scriptsize\ttfamily,
    keywordstyle    = \color{black},
    keywordstyle    = [2] \color{black}, % just to check that it works
    stringstyle     = \color{black},
    commentstyle    = \color{blue}\ttfamily,
    backgroundcolor = \color{backcolour},
    breakatwhitespace=false,
    breaklines=true,
    basewidth=0.55em,
    tabsize=2
}

\usepackage{float}
\usepackage{inconsolata}
\usepackage{xcolor}  % Text colors

\DeclareCaptionLabelFormat{andtable}{#1~#2  \&  \tablename~\thetable}

\graphicspath{{figures/}}

\usepackage{wrapfig}  % for inline figures
\newcommand{\ie}{i.e., }
\newcommand{\eg}{e.g., }

\makeatletter
\renewcommand{\fnum@figure}{\small Figure~\thefigure}  % reduce text size of figure caption
\renewcommand\tablename{\small Table}  % reduce text size of table caption
\makeatother

\usepackage[preprint]{corl_2022} % Uncomment for pre-prints (e.g., arxiv); This is like ``final'', but will remove the CORL footnote.

\title{\algname: Embodied Reasoning\\ through Planning with Language Models}

\author{
  Wenlong Huang${^\dagger}$, Fei Xia${^\dagger}$, Ted Xiao${^\dagger}$, Harris Chan, Jacky Liang, Pete Florence, \\
  \textbf{Andy Zeng, Jonathan Tompson, Igor Mordatch, Yevgen Chebotar, Pierre Sermanet,} \\
  \textbf{Noah Brown, Tomas Jackson, Linda Luu, Sergey Levine, Karol Hausman, Brian Ichter}\\
  Robotics at Google, $^{\dagger}$ equal contribution and alphabetically listed \\
  Project website: \href{https://innermonologue.github.io}{https://innermonologue.github.io}
 
}

\begin{document}
\maketitle

%===============================================================================

\begin{abstract}
Recent works have shown how the reasoning capabilities of Large Language Models (LLMs) can be applied to domains beyond natural language processing, such as planning and interaction for robots.
These embodied problems require an agent to understand many semantic aspects of the world: the repertoire of skills available, how these skills influence the world, and how changes to the world map back to the language.
LLMs planning in embodied environments need to consider not just what skills to do, but also how and when to do them - answers that change over time in response to the agent's own choices.
In this work, we investigate to what extent LLMs used in such embodied contexts can reason over sources of feedback provided through natural language, without any additional training. 
We propose that by leveraging environment feedback, LLMs are able to form an \textit{inner monologue} that allows them to more richly process and plan in robotic control scenarios. 
We investigate a variety of sources of feedback, such as success detection, scene description, and human interaction.
We find that closed-loop language feedback significantly improves high-level instruction completion on three domains, including simulated and real table top rearrangement tasks and long-horizon mobile manipulation tasks in a kitchen environment in the real world. 
\end{abstract}

%===============================================================================
\section{Introduction}
\label{sec:introduction}

Intelligent and flexible embodied interaction requires robots to be able to deploy large repertoires of basic behaviors in appropriate ways, sequence these behaviors as needed for long horizon tasks, and also recognize when to switch to a different approach if a particular behavior or plan is unsuccessful.
High-level planning, perceptual feedback, and low-level control are just a few of the sub-tasks that would need to be seamlessly combined together to perform the sort of reasoning required for an embodied agent, such as a robot, to intelligently act in the world.
While conventionally these challenges have been approached from the perspective of planning (e.g., TAMP~\citep{kaelbling2013integrated}) or hierarchical learning (e.g., HRL~\citep{barto2003recent}), effective high-level reasoning about complex tasks also requires semantic knowledge and understanding of the world.

One of the remarkable observations in recent machine learning research is that large language models (LLMs) can not only generate fluent textual descriptions, but also appear to have rich internalized knowledge about the world~\citep{petroni2019language,jiang2020can,davison2019commonsense,talmor2020olmpics,roberts2020much}. When appropriately conditioned (e.g., prompted), they can even carry out some degree of deduction and respond to questions that appear to require reasoning and inference~\citep{chowdhery2022palm,brown2020language,wei2022chain,kojima2022large,lampinen2022can,nye2021show}.
This raises an intriguing possibility: beyond their ability to interpret natural language instructions,
can language models further serve as reasoning models that combine multiple sources of feedback and become interactive problem solvers for embodied tasks, such as robotic manipulation?

Prior studies show that language helps humans internalize our knowledge and perform complex relational reasoning through \textit{thinking in language}~\citep{vygotsky2012thought,carruthers1998thinking,vygotsky1994tool,vygotsky1967play,colas2022vygotskian}. Imagine the ``inner monologue'' that happens when a person tries to solve some task: ``I have to unlock the door; let me try to pick up the key and put it in the lock... no, wait, it doesn't fit, I'll try another one... that one worked, now I can turn the key.''
The thought process in this case involves choices about the best immediate action to solve the high-level task (``pick up the key''), observations about the outcomes of attempted actions (``it doesn't fit''), and corrective actions that are taken in response to these observations (``I'll try another one'').
Inspired by the human thought process, we propose that such an inner monologue is a natural framework for incorporating feedback for LLMs.

Our work studies these questions by combining LLMs with various sources of textual feedback, only utilizing few-shot prompting without any additional training. 
We observe that similarly to recent work~\citep{zeng2022socratic}, natural language provides a universal and interpretable interface for such grounding of model communication and allows them to incorporate their conclusions in an overarching inner monologue driven by a language model.
While prior work has investigated using language models as planners \cite{huang2022language,ahn2022can} or incorporating multimodal-informed perception through language \cite{zeng2022socratic}, to the best of our knowledge no work has studied the critical link of not only planning with language, but also informing \textit{embodied feedback with language}, which we investigate in this work.

Specifically, we study methods and sources of feedback for closing the agent-environment loop via an inner monologue and their impact on downstream execution success and new capabilities arising from such interaction.
In particular, we combine multiple perception models that perform various tasks such as language-conditioned semantic classification or language-based scene description, together with feedback provided by a human user that the robot is cooperating with.
To execute the commands given by a user, the actions are chosen from a set of pre-trained robotic manipulation skills together with their textual descriptions that can be invoked by a language model.
Our proposed system~\algname chains together these various components (perception models, robotic skills, and human feedback) in a shared language prompt, enabling it to successfully perform user instructions.

%\todo{describe the results and contributions}
Finally, we show that~\algname, without requiring additional training beyond a frozen language model and pre-trained robotic skills, can accomplish complex, long-horizon, and unseen tasks in simulation as well as on two real-world robotic platforms. Notably, we show that it can efficiently retry under observed stochastic failure, replan under systematic infeasibility, or request human feedback for ambiguous queries, resulting in significantly improved performance in dynamical environments. As a demonstration of the versatility of LLMs and grounded closed-loop feedback, we additionally show several surprising capabilities emerging from the inner monologue formulation, including continued adaptation to new instructions, self-proposed goals, interactive scene understanding, multilingual interactions, and more.

\begin{figure}[t]
\centering
\includegraphics[width=1.0\textwidth]{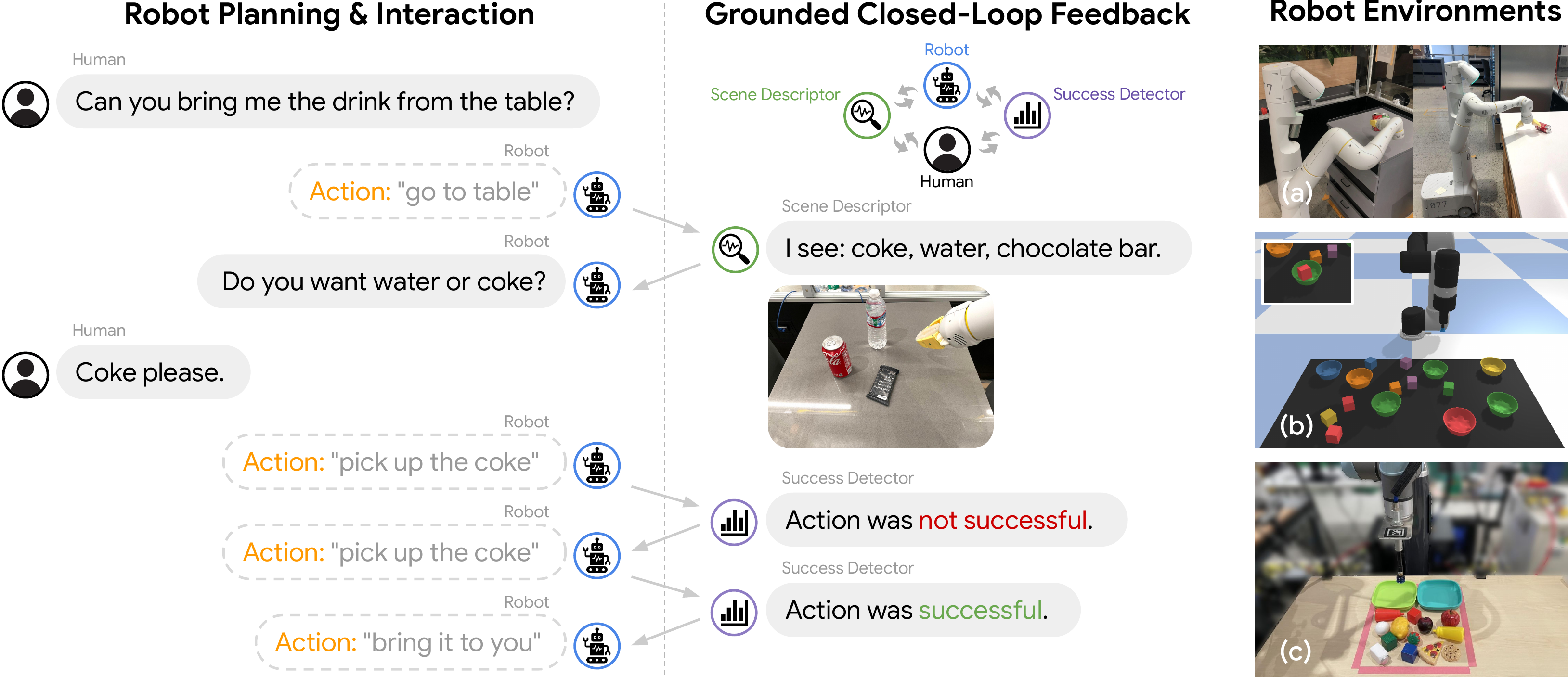}
\caption{\small\textbf{Inner Monologue} enables grounded closed-loop feedback for robot planning with large language models by leveraging a collection of perception models (e.g., scene descriptors and success detectors) in tandem with pretrained language-conditioned robot skills. Experiments show our system can reason and replan to accomplish complex long-horizon tasks for (a) mobile manipulation and (b,c) tabletop manipulation in both simulated and real settings. 
}
\label{fig:sys_diagram}
\vspace{1.4em}
\end{figure}

%===============================================================================
\section{Related Work}
\label{sec:related}
\paragraph{Task and Motion Planning.}
% \textit{(General prior task planning works)}
Task and motion planning \cite{kaelbling2010hierarchical,srivastava2014combined} requires simultaneously solving a high-level, discrete task planning problem \cite{fikes1971strips,sacerdoti1975structure,nau1999shop}, and a low-level, continuous motion planning problem \cite{lavalle2006planning}. 
Traditionally, this problem has been solved through optimization \citep{toussaint2015logic,toussaint2018differentiable} or symbolic reasoning \citep{fikes1971strips,nau1999shop}, but more recently machine learning has been applied to aspects of the problem via learned representations, learned task-primitives, and more~\citep{Eysenbach2019search,xu2018neural,xu2019regression,silver2022inventing,shah2022value,srinivas2018universal,kurutach2018learning,akakzia2021grounding,pirk2020modeling}.
Some works utilize language for planning and grounding~\citep{kollar2010toward,tellex2011understanding,bollini2013interpreting,tellex2014asking,kollar2014grounding,blukis2019learning}.
Others have approached the problem through hierarchical learning~\citep{Nair2020HierarchicalFS,xia2021relmogen,shah2022value,li2020hrl4in,Jiang2019LanguageAA,hafner2022deep, mirchandani2021ella}.
In this work, we leverage pre-trained LLMs and their semantic knowledge, along with trained low-level skills, to find feasible plans.

% \HC{Some works on HRL with language: \citep{jiang2019language,mirchandani2021ella}, manipulation \citep{akakzia2021grounding} discovers abstract goals related to language instructions using intrinsically motivated RL.}  % tompson: Incorporated these.

\paragraph{Task Planning with Language Models.}
Various prior works have explored using language as a space for planning~\citep{jansen2020visually,sharma2021skill,huang2022language,li2022pre,ahn2022can,zeng2022socratic}. Similar to ours are recent task planning approaches that leverage pre-trained autoregressive LLMs to decompose abstract, high-level instructions into a sequence of low-level steps executable by an agent~\citep{huang2022language,ahn2022can} in a zero-shot manner. Specifically,~\citet{huang2022language} prompt GPT-3~\citep{brown2020language} and Codex~\citep{chen2021evaluating} to generate action plans for embodied agents, where each action step is semantically translated to an admissible action with a Sentence-RoBERTa model~\citep{liu2019roberta,reimers2019sentence}. 
% TODO: comment on the autoregressive generation aspect too?
\mbox{SayCan}~\citep{ahn2022can} instead grounds the actions by multiplying each candidate action's probability under FLAN~\citep{wei2021finetuned} with the action's value function, which serves as a proxy for affordance~\citep{shah2022value}. However, both approaches effectively produce the plan while assuming that each proposed step is executed successfully by the agent. As a result, these approaches may not be robust in handling intermediate failures in dynamic environments or with poor lower level policies. We explore in~\algname ways to incorporate grounded feedback from the environment into the LLM as we produce each step in the plan. 

\paragraph{Fusing Vision, Language, and Control in Robotics.}
% Be specific that this is large models on semi-supervised objectives.
%\paragraph{End-to-end Visual Language and Control Models in Robotics.}
Various works have investigated strategies for the challenging problem of fusing vision, language, and control~\citep{paxton2019prospection,stepputtis2020language,blukis2020few,lynch2021language,chen2021joint,mees2022matters,yan2022intra}.
% \citet{lynch2021language} introduced a method to train vision-and-language-conditioned imitation learning on  unstructured robotic experience, and conditionable on free-form language, enabled in part by the use of pretrained LLMs to encode input text. %This work demonstrated some live collaboration where a user can adjust instructions to help the robot achieve tasks (for example ``move back" to get unstuck), albeit as a one-sided conversation.
While pretrained LLMs typically train only on text data, pretrained visual-language models (e.g., CLIP \citep{radford2021learning}) are trained on joint image(s) and corresponding text captions via variants of masked language modeling (MLM) objective \citep{devlin2018bert,lu2019vilbert,wang2021simvlm,suglia2021embodied}, contrastive loss \citep{chen2020big,jain2021mural,radford2021learning} or other supervised objectives\citep{sun2022plate,sener2019zero}. 
%CLIP \citep{radford2021learning} uses contrastive learning to align vision and language representation from millions of image-text pairs from the internet.
CLIP has been employed in several robotics and embodied settings in zero-shot manner \citep{khandelwal2022simple}, or combined with Transporter networks \citep{zeng2020transporter} as in CLIPort \citep{shridhar2022cliport}.
Socratic Models \citep{zeng2022socratic} combines several foundation models (e.g., GPT-3 \citep{brown2020language}, ViLD \citep{gu2021vild}) and language-conditioned policies, using language as the common interface, and demonstrated manipulating objects in a simulated vision-based robotic manipulation environment.
% ~\citet{yan2022intra} also explore using language as an abstraction to form ``intra-agent speech'' for embodied tasks, where an image captioning model is trained using human-labeled captions to facilitate down-stream behavioral learning. In contrast,~\algname leverages \textit{existing} knowledge in frozen LLMs to solve temporally-extended tasks in dynamic environments, by chaining together pre-trained skills and embodied feedback from learned models.

% Others to add:
% \begin{itemize}
    % \item MAC - \citep{yu2022modular} - maybe?
    % \item PlaTe - \citep{sun2022plate} - definite add
    % \item Zero-shot anticipation for instructional activities \citep{sener2019zero} - language model to break down tasks for instruction following.
% \end{itemize}

% \begin{itemize}
%     \item LLMs as planners (Wenlong's, SayCan). Limited to open loop planning from LLMs.
%     \item VLMs, CLIPort
%     \item Socratic models
% \end{itemize}

%===============================================================================
\section{Leveraging Embodied Language Feedback with \algname}
\label{sec:alg}	
%===============================================================================
In this section, we introduce how LLMs %a large language models
can act as interactive problem solvers and incorporate embodied environment observations into grounded planning through a process we refer to as \algname.

\subsection{Problem Statement}
Our setting consists of an embodied robotic agent attempting to perform a high-level natural language instruction $i$.
This robotic agent is only capable of executing short-horizon skills from a library of previously trained policies $\pi_k \in \Pi$ with short language descriptions $\ell_k$, which may be trained with reinforcement learning or behavioral cloning.
The ``planner,'' which is a pretrained LLM~\citep{huang2022language,ahn2022can}, attempts to find a sequence of skills to accomplish the instruction.
To observe the environment, the planner has access to textual feedback $o$ from the environment that can be appended to the instruction or requested by the planner. 
The observation $o$ may be success detection, object detection, scene description, visual-question answering, or even human feedback.
Our work studies to what extent the LLM planner is able to reason over and utilize such feedback to ``close the loop'' with the environment and improve planning.

\subsection{\algname}\label{sec:im}

We formulate an ``inner monologue'' by continually injecting information from the various sources of feedback into the LLM planning language prompts as the robot interacts with the environment.
While LLMs have demonstrated exceptional planning capabilities for embodied control tasks~\citep{huang2022language}, prior works have found it crucial to ground LLM predictions with external components such as affordance functions~\citep{ahn2022can} in order to produce useful plans that are executable by robots. However, LLMs used in this context have thus far remained one-directional -- providing a list of skills, without making corrections or leveraging opportunities to replan accordingly.
In contrast, \algname studies settings where grounded environment feedback is provided directly to the LLM in a closed-loop fashion. This promotes improved LLM reasoning in complex long-horizon settings, even before any external affordance-based grounding methods are applied.

Our analysis assumes textual feedback is provided to the planner, but does not assume a single specific method of fusing LLM planning with low-level robotic control or a specific method of extracting environment feedback into language.
Rather than focusing on a particular algorithmic implementation, our aim is to provide a case study on the value of incorporating different types of feedback into closed-loop LLM-based planning.
Thus, \algname in Sec~\ref{sec:result} utilizes language feedback within separate systems that incorporate different LLMs, different methods of fusing planning with control, different environments and tasks, and different methods of acquiring control policies.
We note that in our specific implementations of \algname, we use pre-trained LLMs for planning that are not finetuned, but rather evaluated solely with few-shot prompting; the full prompts can be found in the Appendix.

\subsection{Sources of Feedback}\label{subsec:feedbacks}

\begin{figure}
    \centering
    \includegraphics[width=0.85\linewidth]{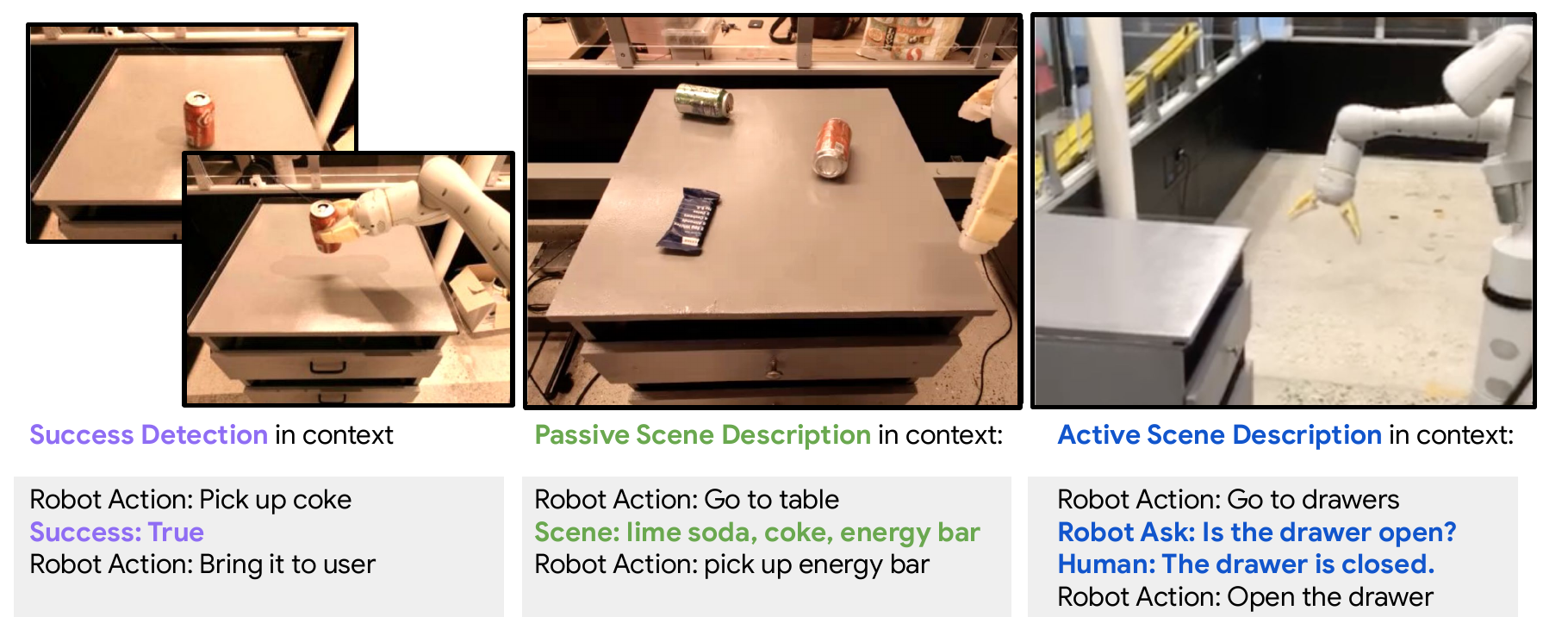}
    \caption{Various types of textual feedback. \textbf{{\color{success_detection_color}Success Detection}} gives task-specific task completion information, \textbf{{\color{scene_description_color}Passive Scene Description}} gives structured semantic scene information at every planning step, and \textbf{{\color{vqa_color}Active Scene Description}} gives unstructured semantic information only when queried by the LLM planner.}
    \label{fig:text_feedback}
    \vspace{1.4em}
\end{figure}

In theory any type of environment feedback can inform the LLM planner, as long as it can be expressed through language.
We focus on the specific forms of feedback shown in Fig~\ref{fig:text_feedback}, which can be broken down into task-specific feedback, such as success detection, and scene-specific feedback (either ``passive'' or ``active''), which describes the scene. Specific instantiations and implementation details of each type of feedback can be found in Sec~\ref{sec:tabletop-experiments}, Sec~\ref{sec:real-tabletop}, and Sec~\ref{sec:mobile-experiments} respectively for each domain.

\paragraph{Success Detection.}
Semantic success detection is a binary classification problem of whether the low-level skill $\pi_k$ has succeeded.
Engineered success detectors can operate on ground-truth state in simulation, while learned success detectors can be trained on real examples of successes and failures in the real world~\citep{lenz2015deep,chu2018real,kalashnikov2021mt,migimatsu2021grounding,cui2022can}.
We use the output of success detectors in language form, which we refer to as \textit{Success} feedback.

\paragraph{Passive Scene Description.}

While there are many ways to describe the semantics contained within a scene, we use the term Passive Scene Description to broadly describe sources of scene feedback that are consistently provided and follow some structure.
Passive Scene Description covers all sources of environment grounding feedback that are automatically provided and injected into the LLM prompt without any active prompting or querying by the LLM planner. One common type of such feedback is object recognition~\cite{liang2015recurrent,ren2015faster,zou2019object,bochkovskiy2020yolov4} -- we refer to the textual outputs of such object recognizers as \textit{Object} feedback. We also demonstrate the use of a task-progress scene description in the simulated tabletop rearrangement environment, to which we refer as \textit{Scene} feedback.

\paragraph{Active Scene Description.}
As the proactive counterpart to Passive Scene Description, Active Scene Description encompasses sources of feedback that are provided directly in response to active queries by the LLM planner.
In this case, the LLM can directly ask a question about the scene, and this question can be answered either by a person, or by another pretrained model, such as a Visual Question Answering (VQA) model~\cite{antol2015vqa,zhou2020unified,cai2020once,alayrac2022flamingo}.
While the previous types of feedback are strictly structured and narrow in their scope, in the Active Scene Description setting the LLM can receive unstructured answers to open-ended questions, allowing it to actively gather information relevant to the scene, the task, or even preferences of the user (in the case of human-provided response).
The combined output we send to the LLM planner includes both the LLM-generated question along with the response.
As we aim to investigate \textit{whether} and \textit{how} a LLM planner can incorporate such feedback and wish to study both structured VQA-style human feedback as well as unstructured human preferences feedback, we only consider human-provided response in this work, which we refer to as \textit{Human} feedback.

\section{Experimental Results}
\label{sec:result}

In order to study how different sources of environment feedback can support a rich inner monologue that enables complex robotic control, we analyze diverse long-horizon manipulation and navigation tasks in simulation and in the real world.
As \algname is not dependent on a specific LLM or a type of grounding feedback, we study different \algname implementations in three environments with different LLM planning methods and different sources of feedback from the environment.
Below, we show results for a tabletop manipulation environment in sim (Sec~\ref{sec:tabletop-experiments}) and real (Sec~\ref{sec:real-tabletop}) as well as a mobile manipulation environment in real (Sec~\ref{sec:mobile-experiments}). For more details about the experiment setup and results, please refer to the Appendix.

\begin{figure}[t]
\centering
\includegraphics[width=1.0\linewidth]{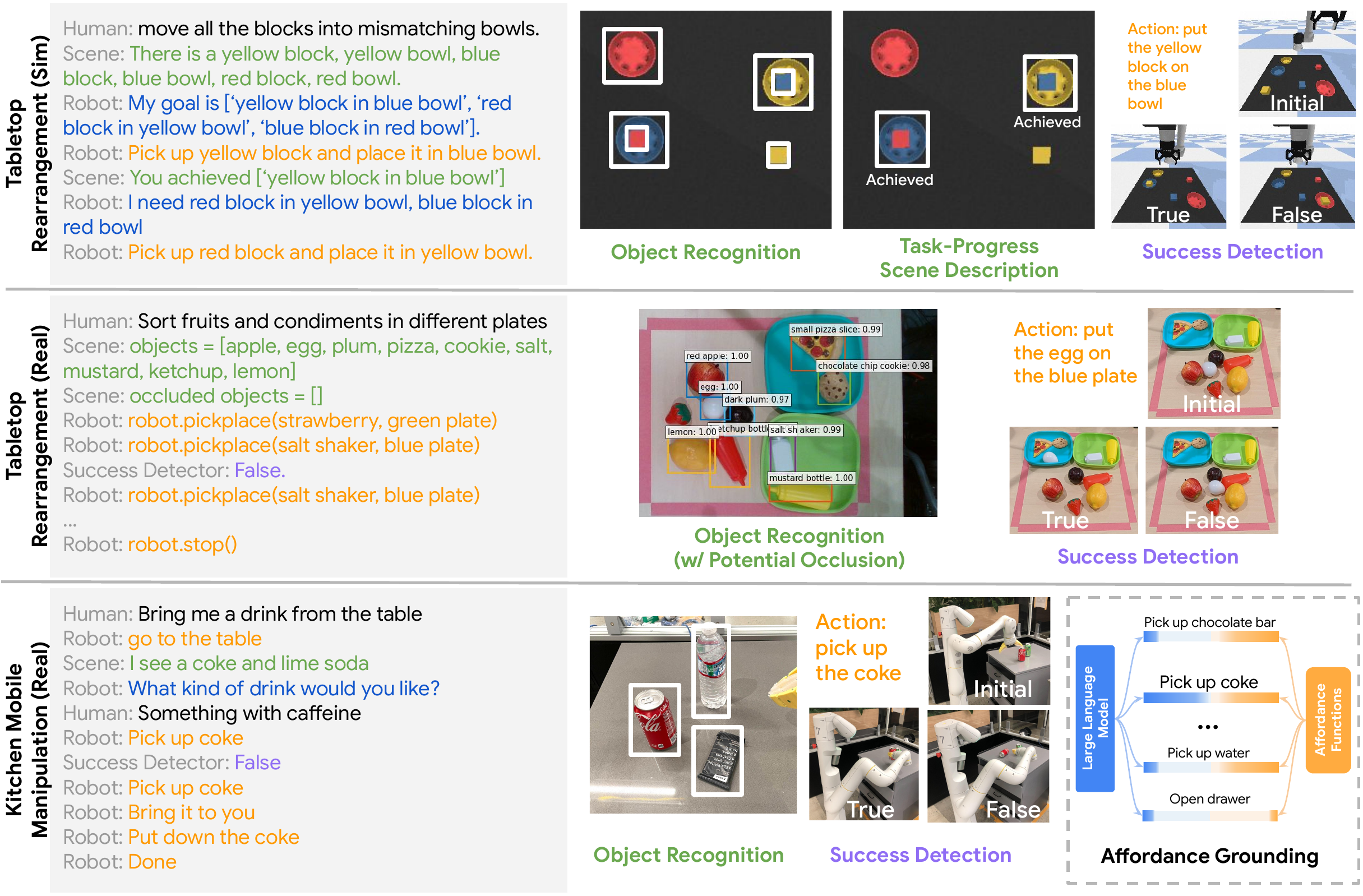}
\caption{\small Different instantiations of~\algname in three distinct domains -- simulated tabletop rearrangement (\textit{top}), real-world tabletop rearrangement (\textit{middle}), and real-world kitchen mobile manipulation (\textit{bottom}). Each domain uses different prompts and different feedback models. Sharing across the domains is the same~\algname formulation that uses a pre-trained langauge model to take in a \textbf{human instruction} and decompose it into a sequence of \textbf{\color{robot_action_color}{actionable steps}} by the agent, while accounting for injected embodied feedback from different models, such as \textbf{\color{scene_description_color}{object recognizers}} and \textbf{\color{success_detection_color}{success detectors}}. In real-world kitchen mobile manipulation domain (\textit{bottom}), we additionally ground the actions using pre-trained affordance functions built in~\citep{ahn2022can}, which do not communicate back to the language model.}
\label{fig:all_domains}
\vspace{-2.0em}
\end{figure}

\subsection{Simulated Tabletop Rearrangement}\label{sec:tabletop-experiments}

We experiment with vision-based block manipulation tasks in a Ravens-based~\citep{zeng2020transporter} simulation environment to evaluate our method against several baselines and ablate across varying amounts of textual feedback. Given a number of blocks and bowls on a table, a robotic arm containing a gripper is tasked with rearranging these objects in some desired configuration, specified by natural language (e.g., ``putting the blocks in the bowls with matching colors''). We evaluate each method on four seen tasks and four unseen tasks, where seen tasks may be used for training (in the case of supervised baseline) or used as few-shot prompting for LLM planner.

This instantiation of~\algname uses (i) InstructGPT~\citep{brown2020language,ouyang2022training} as the LLM for multi-step planning~\citep{huang2022language,ahn2022can}, (ii) scripted modules to provide language feedback in the form of object recognition (\textit{Object}), success detection (\textit{Success}), and task-orogress scene description (\textit{Scene}), and (iii) a pre-trained language-conditioned pick-and-place primitive (similar to CLIPort~\citep{shridhar2022cliport} and Transporter Nets~\citep{zeng2020transporter}).
\textit{Object} feedback informs the LLM planner about the objects present in the scene, and the variant using only \textit{Object} feedback is similar to the demonstrated example in~\citep{zeng2022socratic} in this environment. \textit{Success} feedback informs the planner about success/failure of the most recent action. However, in the presence of many objects and test-time disturbances, the complex combinatorial state space requires the planner to additionally reason about the overall task progress (e.g., if the goal is to stack multiple blocks, the unfinished tower of blocks may be knocked over by the robot). Therefore, task-progress scene description (\textit{Scene}) describes the semantic sub-goals inferred by the LLM towards completing the high-level instruction that are achieved by the agent so far. For the variant that uses \textit{Object + Scene} feedback, due to the additional reasoning complexity, we find that adding chain-of-thought~\cite{wei2022chain, lampinen2022can,nye2021show} can improve the consistency between inferred goals and achieved goals.

We additionally compare to a multi-task CLIPort policy directly trained on long-horizon task instructions (i.e., without using LLM for planning). Because CLIPort is a single-step policy and does not terminate spontaneously during policy rollout, we report CLIPort evaluations with oracle termination (i.e., repeat until oracle indicates task completion) and fixed-step termination (i.e., repeat for $k$ steps). While~\algname terminates when the LLM stops generating new steps, we similarly set the maximum number of steps to be $k$ for practical considerations. We use $k=15$. To simulate real-world disturbances and evaluate the system's robustness to disturbances, we add Gaussian noise to multiple levels of the system at test time: $\mathcal{N}(0, 3)$ for pixel observation, $\mathcal{N}(0, 2.5)$ for policy primitive (i.e., pick-place pixel heatmaps), $\mathcal{N}(0, 0.02m)$ for place locations.

Results are shown in Table~\ref{tab:ravens}, and example prompt can be found in Fig~\ref{fig:all_domains}.

\vspace{-0.5em}
\begin{table}[H]
\setlength\tabcolsep{3.8pt}
\begin{center}
\newfootnotesize  % newfootnotesize is between footnotesize and scriptsize
\renewcommand{\arraystretch}{1.05}
\rowcolors{1}{white}{white}
\begin{tabular}{@{}clccccc@{}}
\toprule
& &  &  & +\textbf{LLM} & \multicolumn{2}{c}{+\textbf{\algname}}  \\
\cmidrule(lr){5-5} \cmidrule(lr){6-7}
& \textbf{Tasks} & \textbf{CLIPort} & \textcolor{gray}{+oracle} & \textit{Object} & \textit{Object + Success} & \textit{Object + Scene} \\
\midrule
& ``Pick and place'' & 24.0\% & \textcolor{gray}{74.0\%} & 80.0\%  & 90.0\% & \textbf{94.0\%} \\
& ``Stack all the blocks'' & 2.0\% & \textcolor{gray}{32.0\%} & 4.0\% & 10.0\% & \textbf{26.0\%} \\
\textbf{Seen Tasks } & ``Put all the blocks on the [x] corner/side'' & 2.0\% & \textcolor{gray}{32.0\%} & \textbf{30.0\%} & 28.0\% & \textbf{30.0\%} \\
&  ``Put all the blocks in the [x] bowl'' & 32.0\% & \textcolor{gray}{94.0\%} & 52.0\%  & 46.0\% & \textbf{56.0\%} \\
\midrule
& ``Put all the blocks in different corners'' & 0.0\% & \textcolor{gray}{0.0\%} & 20.0\% & 20.0\% & \textbf{26.0\%} \\
& ``Put the blocks in their matching bowls'' & 0.0\% & \textcolor{gray}{0.0\%} & 56.0\% & 70.0\% & \textbf{82.0\%} \\
\textbf{Unseen Tasks} & ``Put the blocks on mismatched bowls'' & 0.0\% & \textcolor{gray}{0.0\%} & 62.0\% & 76.0\% & \textbf{86.0\%} \\
& ``Stack all the blocks on the [x] corner/side'' & 0.0\% & \textcolor{gray}{0.0\%} & 0.0\% & 4.0\% & \textbf{6.0\%} \\
\bottomrule
\end{tabular}
\caption{Success rates for various methods, averaged across 50 episodes in Ravens-based environment with test-time disturbances. CLIPort + oracle indicates that CLIPort was provided a ``termination'' oracle. Although CLIPort can receive visual feedback from the environment, we show that LLM-informed feedback can effectively enable the planner to retry/replan in the presence of failures, while enjoying the generalization benefits of LLMs to unseen tasks.
}
\label{tab:ravens}
\end{center}
\end{table}
\vspace{-1.8em}

\paragraph{Analysis.} As shown in Table~\ref{tab:ravens}, all variants of~\algname perform well on seen tasks, and by leveraging rich semantic knowledge in the pre-trained LLM, the performance can be directly translated to unseen tasks without further training. Furthermore, aligned with our hypothesis,~\algname with \textit{Object + Scene} performs the best because of its ability to keep track of all goal conditions and currently achieved goals.
Finally, we observe that non-hierarchical and solitary systems such as CLIPort (i) struggle at generalizing to unseen long-horizon tasks under test-time disturbances, and (ii) on training tasks, an oracle is also often required to indicate task completion for good performance.

\subsection{Real-World Tabletop Rearrangement}\label{sec:real-tabletop}
We evaluate~\algname on a real-world robot platform designed to resemble the simulation experiments presented in Sec~\ref{sec:tabletop-experiments} using motion primitives for tabletop pick and place. The setup consists of a UR5e robot arm equipped with a wrist-mounted Intel RealSense RGB-D camera overlooking a workspace of diverse objects -- from toy blocks to food items to condiments (shown in Fig~\ref{fig:all_domains}). We use an instantiation of~\algname using (i) InstructGPT~\citep{brown2020language,ouyang2022training} as the LLM for multi-step planning, (ii) a pretrained open-vocabulary object recognition with MDETR \cite{kamath2021mdetr} to generate a list of currently visible objects and list of previously visible objects that are no longer visible (\textit{Object}), (iii) heuristics on the object bounding box predictions from MDETR for Success Detection (\textit{Success}), and (iv) a zero-shot pick and place policy that uses an LLM to parse target objects from a language command (\eg given by the planner) and then executes scripted suction-based picking and placing primitives at the center of the target objects' bounding boxes. Aside from the pretraining of the LLM and MDETR (which are available out-of-the-box), the system does not require any model finetuning to perform pick and place tasks with new objects.

We investigate two tasks: (i) a simple 3-block stacking task where 2 blocks are already pre-stacked, and (ii) a more complex long-horizon sorting task to place food in one plate and condiments in another (where categorizing food versus condiments is autonomously done by the LLM planner). Since default pick and place performance is generally quite high on the system, we artificially inject Gaussian noise into the policy actions (\ie add standard deviation $\sigma{=}4\text{mm}$ clipped at $2\sigma$) to stress test recovery from failures via replanning with grounded closed-loop feedback. Results are presented in Table~\ref{tab:real-ur5}. Note that the system is also subject to noisy object and success detections due to the additional challenge of real-world perception and clutter.

\begin{minipage}{\textwidth}
    \centering
    \small
    \setlength\tabcolsep{4.8pt}
    \begin{tabular}{@{}lcccc@{}}
        \toprule
        & \textbf{LLM}  & \multicolumn{3}{c}{+\textbf{\algname}}  \\
        \cmidrule(lr){2-2} \cmidrule(lr){3-5}
        \textbf{Task Family} & \textit{Object} & \textit{Object} &  \textit{Success} & \textit{Object + Success} \\
         \midrule
        Finish 3-block stacking & 20\% & 40\% & 40\% & \textbf{100\%}  \\
        Sort fruits from bottles & 20\% & 50\% & 40\% & \textbf{80\%} \\
        \textbf{Total} & 20\% & 45\% & 40\% & \textbf{90\%} \\
        \bottomrule
    \end{tabular}
    \vspace{0.5em}
    \captionof{table}{\algname (with object recognition and success detection feedback) on a real pick and place robot exceeds the performance of baseline alternatives, as measured by average task success rates over 10 runs. For both tasks, we observe significant improvement in~\algname compared to the baseline, with the two embodied feedback being complementary to each other.}\label{tab:real-ur5}
    %\end{minipage}
  \end{minipage}
\vspace{-1.0em}

\paragraph{Analysis.} We compare different variants of~\algname with different LLM-informed closed-loop feedback, as well as an open-loop variant that only runs object recognition once at the beginning of the task (similar to the system demonstrated in \cite{zeng2022socratic}). The partial 3-block stacking task highlights an immediate failure mode of this baseline, where the initial scene description struggles to capture a complete representation of the scene (due to clutter and occlusion) to provide as input to the multi-step planner. As a result, the system only executes one pick and place action -- and cannot recover from mistakes. To address these shortcomings,~\algname (\textit{Object + Success}) leverages closed-loop scene description and success detection after each step, which allows it to successfully replan and recover from policy mistakes.

Additional ablations with~\algname also show that (i) common failures induced by lack of closed-loop scene description are largely due to initially occluded objects not being part of the LLM generated plans, and (ii) failures induced by lack of success detection come from not retrying pick and place actions that have failed due to policy noise. Overall, we observe that both components are complementary and important in maintaining robust recovery modes for real rearrangement tasks.

\subsection{Real-World Mobile Manipulator in a Kitchen Setting}\label{sec:mobile-experiments}

We implement \algname in a robotic system using the kitchen environment and task definitions described in SayCan \citep{ahn2022can}. The \href{https://everydayrobots.com/}{Everyday Robots} robot, a mobile manipulator with RGB observations, is placed in an office kitchen to interact with common objects using concurrent~\citep{xiao2020nonblocking} continuous closed-loop control.

The baseline, SayCan \citep{ahn2022can}, is a method that plans and acts in diverse real world scenarios by combining an LLM with value functions of underlying control policies.
While SayCan creates plans that are grounded by the affordances of value functions, the LLM predictions in isolation are never given any closed-loop feedback.

We use an instantiation of \algname that uses (i) PALM~\citep{chowdhery2022palm} as the LLM for multi-step planning, (ii) value functions from pre-trained control policies for affordance grounding~\citep{ahn2022can}, (iii) a learned visual classification model for success detection feedback (\textit{Success}), (iv) human-provided object recognition feedback (\textit{Object}), and (v) pre-trained control policies for relevant skills in the scene.
We also perform a case study where we allow the agent to ask questions and source \textit{Human} feedback directly; results are shown in Fig~\ref{fig:emergent_adapt} and the Appendix.

We evaluate methods on 120 evaluations over three task families: 1) four manipulation tasks, 2) two dexterous manipulation tasks utilizing drawers, and 3) two long-horizon combined manipulation and navigation tasks.
In order to better study how \algname improves reasoning in especially challenging scenarios, we consider an experiment variant where we add adversarial disturbances during control policy executions that cause skill policy rollouts to fail.
These disturbances may be fairly simple and just require the policy to try again, or it may complex enough that the policy needs to replan and select a completely new skill.
While these failures occur naturally even without perturbances, the adversarial disturbances creates a consistent comparison between methods that requires retrying or replanning to accomplish the original instruction.

\begin{minipage}{\textwidth}
   \vspace{3mm}
  \begin{minipage}[]{0.55\textwidth}
    % \vspace{-1.5em}
    \centering
    \small
    % \newfootnotesize
    \begin{tabular}{@{}lccc@{}}
    \toprule
    % \cline{3-6}
    %\hhline{~---}
    &  & \multicolumn{2}{c}{+\textbf{\algname}}  \\
    \cmidrule(lr){3-4}
    \textbf{Task Family} & \textbf{SayCan} & \textit{Success} &  \textit{Object + Success} \\
     \midrule
     \textbf{\textcolor{gray}{\small No Disturbances }} &  &  &  \\
    Manipulation & 50.0\% & 62.5\% & \textbf{75.0\%} \\
    Mobile Manipulation & 50.0\% & 50.0\% & \textbf{75.0\%} \\
    Drawers & 83.3\% & 83.3\% & \textbf{100.0\%}\\
    \midrule
    \textbf{\textcolor{gray}{\small With Disturbances }} &  &  &  \\
    Manipulation & 12.5\% & 25.0\% & \textbf{33.3\%} \\
    Mobile Manipulation & 0.0\% & 25.0\% & \textbf{75.0\%}  \\
    Drawers & 0.0\% & \bf{44.4\%} & \bf{44.4\%}\\
    \midrule
    Total & 30.8\%	 & 48.7\% & \textbf{60.4\%} \\
    \bottomrule
    \end{tabular}
    \vspace{0.5em}
    \captionof{table}{Averaged success rate across 120 evaluations on several task families in our real-world mobile manipulation environment. We consider a standard setting and adversarial setting with external human disturbances. In all cases, LLM-informed embodied feedback is shown to be effective in improving robustness of the system, especially when low-level policies are prone to failures.}\label{tab:saycan}
  \end{minipage}
  \hfill
  \begin{minipage}[]{0.43\textwidth}
      \vspace{-4mm}
    \centering
    \includegraphics[width=1.0\textwidth]{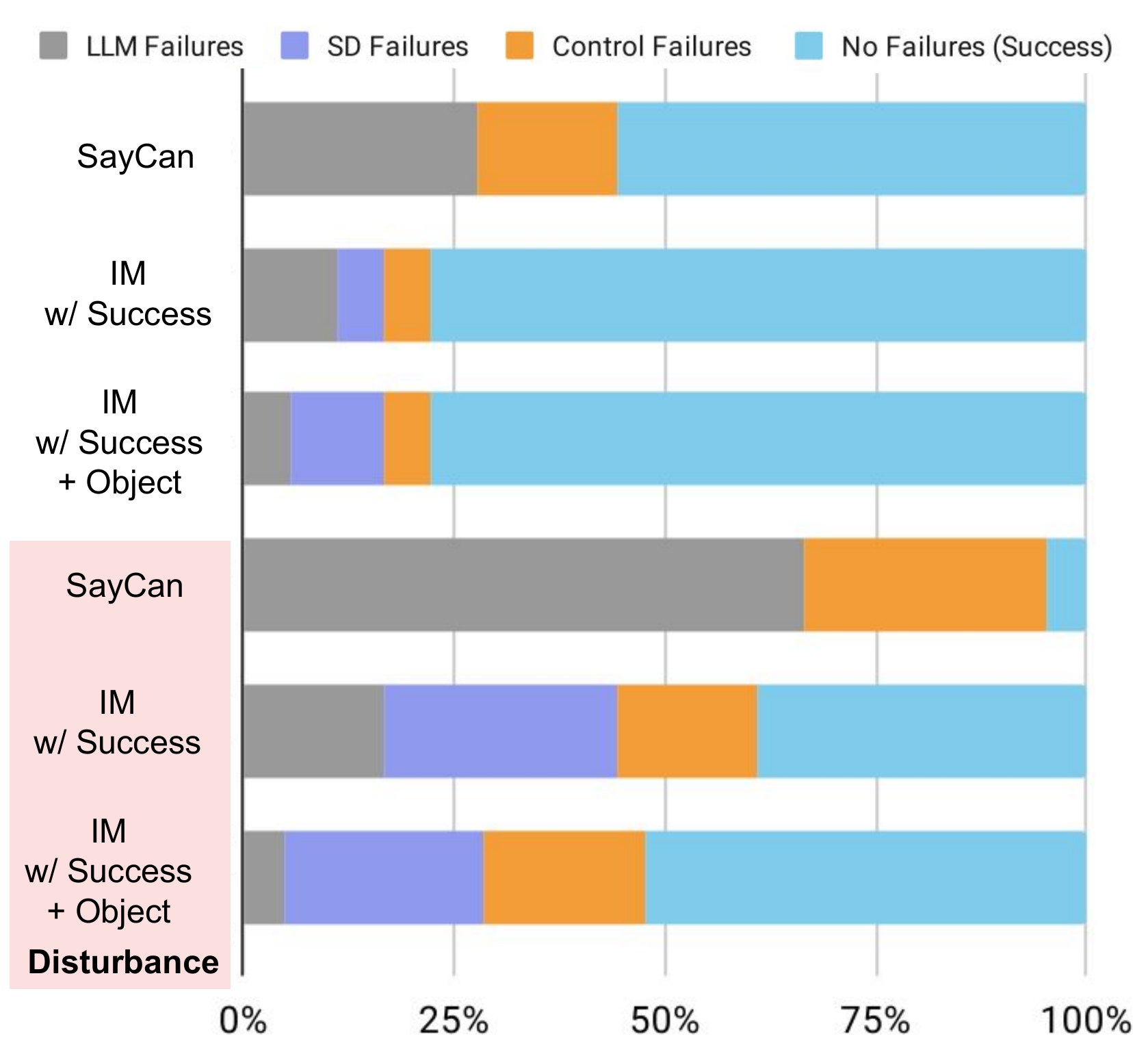}
    \vspace{-4mm}
    \captionof{figure}{Failure causes on 120 evaluations. When disturbances are added (red), only the \algname variants consistently complete the instructions.}\label{fig:failures}
  \end{minipage}
  \end{minipage}

\paragraph{Analysis.} The results of real robot experiments are shown in Table~\ref{tab:saycan}. Without adversarial disturbances, the baseline method SayCan performs reasonably on all three task families, yet incorporating LLM-informed feedback on skill success/failure and presence of objects allows the~\algname to effectively retry or replan under natural failures, providing further improvement to the baseline. The most notable difference is in the cases with adversarial disturbances when a policy failure is forced to occur. Without any LLM-informed feedback SayCan has success rate close to 0\% since it is does not have explicitly high-level retry behavior. \algname significantly outperforms SayCan because of its ability to invoke appropriate recovery modes depending on the environment feedback. In-depth analysis on the failure causes indicates that \textit{Success} and \textit{Object} feedback can effectively reduce LLM planning failures and thus overall failure rate, albeit at the cost of introducing new failure modes to the system.

\subsection{Emergent Capabilities}
\label{sec:emergent}
Although LLMs can generate fluent continuation from the prompted examples, we surprisingly find that, when informed with environment feedback,~\algname demonstrates many impressive reasoning and replanning behaviors beyond the examples given in the prompt. Using a pre-trained LLM as the backbone, the method also inherits many of the appealing properties from its versatility and general-purpose language understanding. In this section, we demonstrate a few of these emergent capabilities.

\begin{figure}[t]
\centering

\begin{subfigure}{.52\textwidth}
  \centering
  \includegraphics[width=1.0\linewidth]{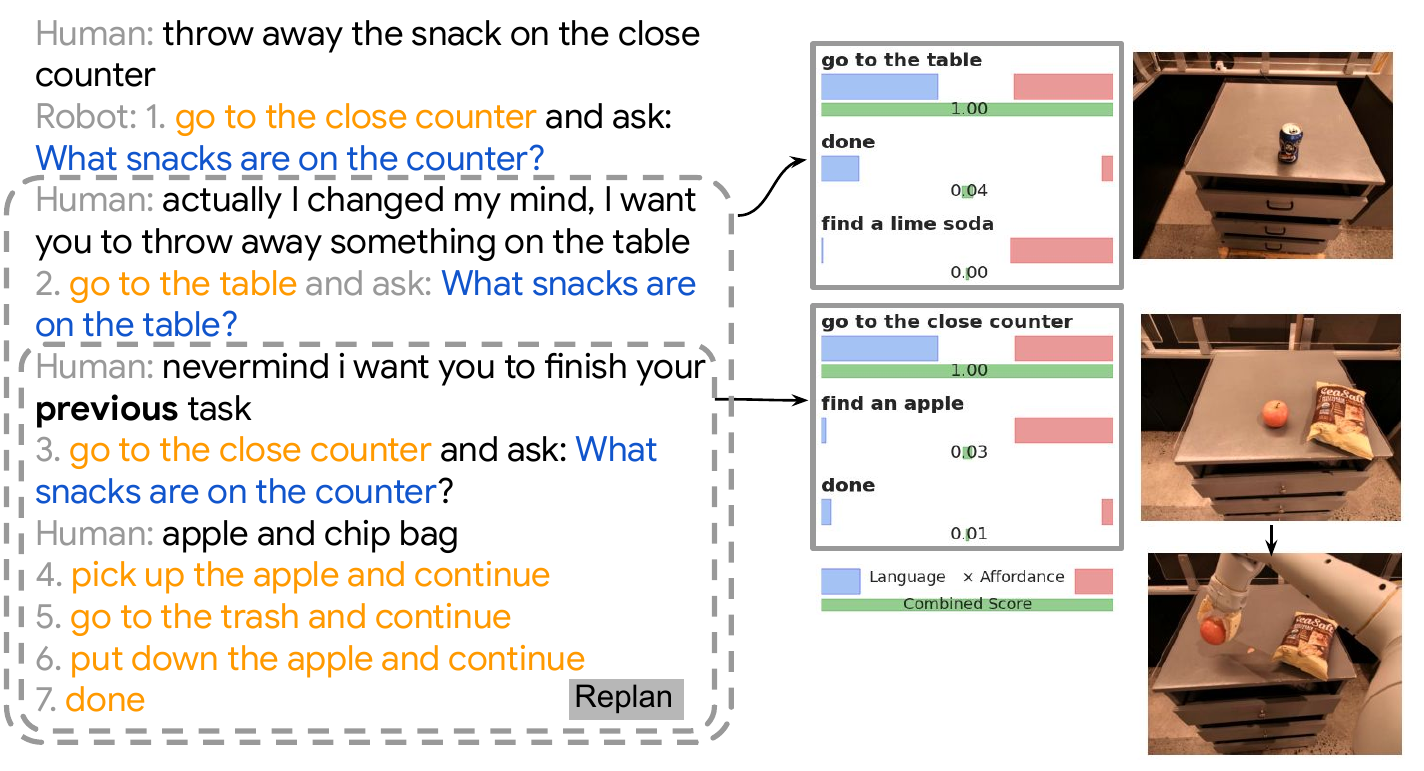}
  \caption{Continued Adaptation to New Instructions}
  \label{fig:emergent_adapt}
\end{subfigure}
\begin{subfigure}{.46\textwidth}
  \centering
  \includegraphics[width=1.0\linewidth]{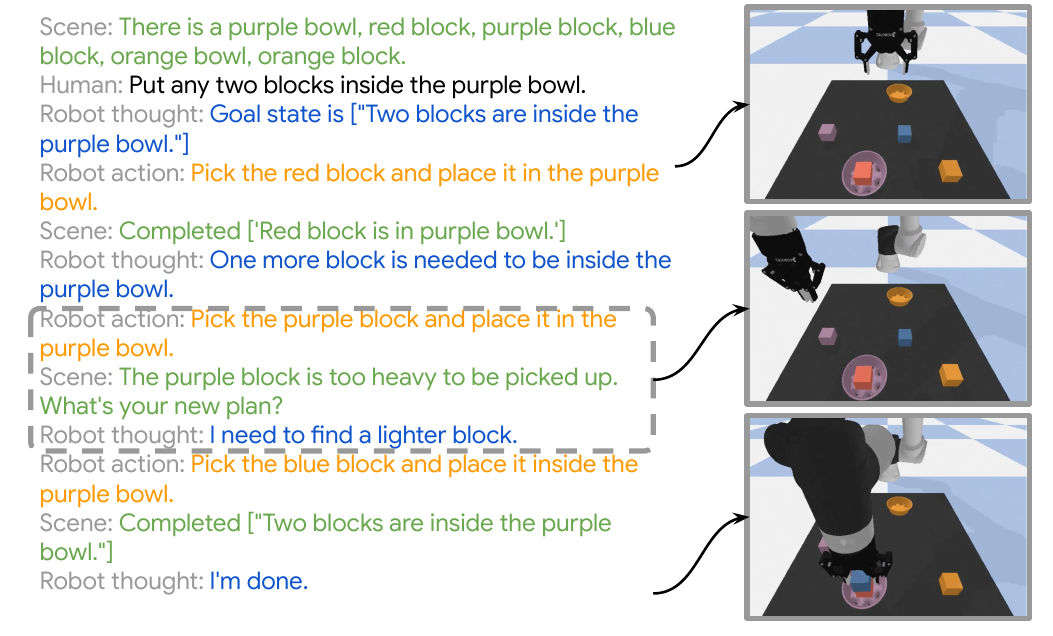}
  \caption{Self-Proposing Goals under Infeasibility}
  \label{fig:emergent_spontaneous}
\end{subfigure}
\begin{subfigure}{.49\textwidth}
  \centering
  \includegraphics[width=1.0\linewidth]{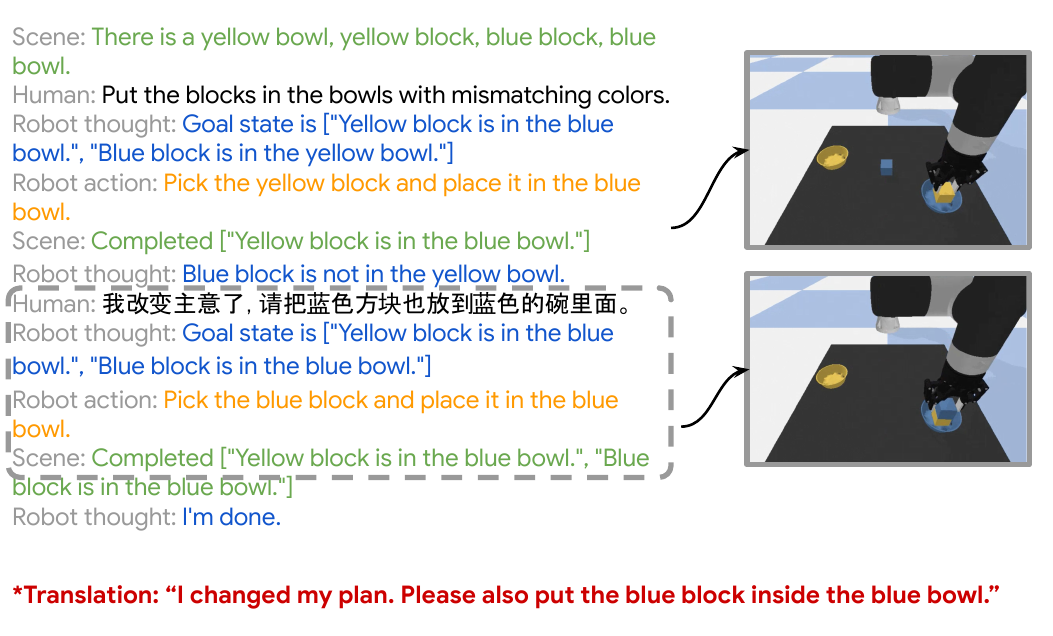}
  \caption{Multilingual Interaction}
  \label{fig:emergent_multilingual}
\end{subfigure}
\begin{subfigure}{.49\textwidth}
  \centering
  \includegraphics[width=1.0\linewidth]{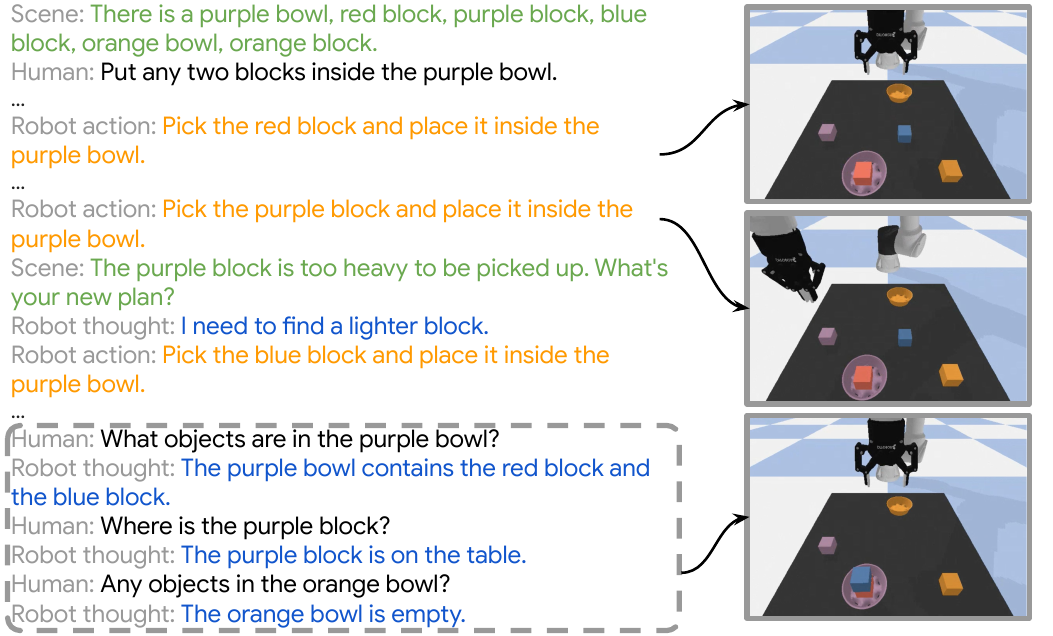}
  \caption{Interactive Scene Understanding}
  \label{fig:emergent_scene}
\end{subfigure}
\caption{Informing LLM with embodied feedback enables many emergent capabilities, all of which are achieved without similar prompted examples. For instance,~\algname can continually adapt to new instructions given by humans, propose new goals to achieve when faced with infeasibility for the previous plan, interact with humans in different natural languages, and answer questions about the current scene given past actions and feedback.}
\label{fig:emergent}
\vspace{1.5em}
\end{figure}

\paragraph{Continued Adaptation to New Instructions.}
Although not explicitly prompted, the LLM planner can react to human interaction that changes the high-level goal mid-task. Fig~\ref{fig:emergent_adapt} demonstrates a challenging case, where \textit{Human} feedback changes the goal during the plan execution, and then changes the goal yet again by saying ``finish the previous task". We can see that the planner incorporates the feedback correctly by switching tasks twice. In another instance, despite not being explicitly prompted to terminate after a human says ``please stop’’, the LLM planner generalizes to this scenario and predicts a ``done’’ action.

\paragraph{Self-Proposing Goals under Infeasibility.}
Instead of mindlessly following human-given instructions,~\algname can also act as an interactive problem solver by proposing alternative goals to achieve when the previous goal becomes infeasible. In Fig~\ref{fig:emergent_spontaneous}, to solve the task ``put any two blocks inside the purple bowl'',~\algname first attempts an action of picking up the purple block -- the action fails as the purple block is intentionally made to be too heavy for the robot. After a hint ``the purple block is too heavy", it proposes to ``find a lighter block'' and successfully solves the task in the end.

\paragraph{Multilingual Interaction.}
Pre-trained LLMs are known to be able to translate from one language to another, without any finetuning. We observe that such multilingual understanding also transfers to the embodied settings considered in this work. Specifically, in Fig~\ref{fig:emergent_multilingual}, the human-provided new instruction is written in Chinese, but the LLM can correctly interpret it, re-narrate it as a concrete goal to execute in English, and accordingly replan its future actions. Occasionally, we find that this capability even extends to symbols and emojis.

\paragraph{Interactive Scene Understanding.}
We also observe that~\algname demonstrates interactive understanding of the scene using the past actions and environment feedback as context. In Fig~\ref{fig:emergent_scene}, after a task instruction has been executed, we turn to ask questions about the scene, again a structure that has not appeared in the prompt. Surprisingly, we find that it can often correctly answer these questions that require temporal and embodied reasoning.

\paragraph{Robustness to Feedback Order.}
In the main experiments of the paper, we prompted the language model following certain conventions. For instance, in the simulated tabletop domain, the convention is \texttt{[Robot action, Scene, and Robot thought]}. In practice, we find that the LLM planner is robust to occasionally swapping the order of feedback. In Appendix Fig~\ref{fig:emergent_order}, a new human instruction is injected in the middle of the plan execution, but this structure has not been seen in the example prompts. Yet the planner recognizes the change and generates a new ``Robot thought: Goal state is…’’ statement allowing it to solve the new task.
\paragraph{Robustness to Typos.}
Inherited from the LLM backbone, our approach is robust to typos in human instruction, as seen in Appendix Fig~\ref{fig:emergent_typos}.

Despite the appealing findings about these emergent capabilities, we observe that they are of varying levels of consistency when no similar examples have been provided in the prompt, likely limited by the current capabilities of the language models. However, we believe that further investigations into these behaviors and addressing their limitations would each lead to exciting future directions.

%===============================================================================
\vspace{-0.2cm}
\section{Limitations}
\vspace{-0.2cm}

\label{sec:limitations}
\paragraph{Limiting assumptions and failure modes.} In Sec~\ref{sec:tabletop-experiments} and Sec~\ref{sec:mobile-experiments}, we assume access to oracle scene descriptors in the form of human observers or scripted systems to provide textual description back to the LLM planner.
We study the viability of learned systems scene desription and object recognition in Appendix Table~\ref{tab:saycan_object_detection}.
As for failure modes, \algname may fail due to several sources of errors: (1) success detections, (2) LLM planning errors, and (3) control errors.
False negative predictions from the success detector lead to additional retry attempts, while false positive predictions add adversarial partial observability to the environment.
In some instances, we found that the LLM planners ignored the environment feedback and still proposed policy skills involving objects not present in the scene. 

\paragraph{Limitations of results and future work.}
The performance of low-level control policies limits not only overall high-level instruction completion performance, but also limits the scope of tasks that the LLM is able to reason over: no matter how much the LLM reasoning improves, it can still be bottlenecked by what low-level control policies are able to achieve.
Future work remains open on several fronts.
First, with advances in image/video captioning and visual-question answering, a fully automated system of \algname can be implemented without a human in the loop as an oracle. 
Second, improvements can be made on how to aggregate potentially inaccurate sources of information, such as using text to describe the uncertainty of the feedback modules, or including additional feedback modules for safety and ethics for the proposed plans. 

%===============================================================================
\section{Conclusion}
\label{sec:conclusion}
In this work, we investigated the role that environment feedback plays for LLMs reasoning in tasks involving embodied robotic planning and interaction.
We presented a general formulation \algname that combines different sources of environment feedback with methods fusing LLM planning with robotic control policies.
Then, we studied how these methods scaled to three different robotic manipulation settings in simulation and the real world.
We found that environment feedback significantly improves high-level instruction completion, especially in challenging scenarios with adversarial disturbances.
Finally, we analyze emergent capabilities of \algname that highlight how closed-loop language feedback enables replanning even in complex unseen settings.

%===============================================================================

% 
% The acknowledgments are automatically included only in the final and preprint versions of the paper.
\acknowledgments{The authors would like to thank Kanishka Rao and Vincent Vanhoucke for valuable feedback and discussions. In addition, the authors would like to acknowledge the large team who built \citep{ahn2022can}, upon which we construct our Kitchen Mobile Manipulation experiments.}
\clearpage
%===============================================================================

\bibliography{ref}  % .bib

\newpage
\appendix
\section{\algname Implementation Details}\label{sec:app_policies}

We study three different implementations of \algname for each of the experimental settings.
While each version incorporates textual environment feedback into planning, there are differences in the internal components of each system, as seen in Table~\ref{tab:implementation_comparison}.

\begin{table}[H]
\setlength\tabcolsep{3.8pt}
\begin{center}
\newfootnotesize  % newfootnotesize is between footnotesize and scriptsize
\renewcommand{\arraystretch}{1.05}
\rowcolors{1}{white}{white}
\begin{tabular}{@{}clll@{}}
\toprule
& \textbf{Tabletop Rearrangement (Sim)} & \textbf{Tabletop Rearrangement (Real)} & \textbf{Kitchen Mobile Manipulation (Real)} \\ % & \textbf{+ Success \& Scene} \\
\midrule
\textbf{LLM} & InstructGPT-1.3B & InstructGPT-1.3B & PALM-540B \\ 
\midrule
\textbf{Policies} & CLIPort + TransporterNets & MDETR + heuristics & Learned policies \\
\midrule
\textbf{\textit{Success} Feedback} & Ground truth + heuristics & MDETR + heuristics & Finetuned CLIP \\
\midrule
\textbf{\textit{Object} Feedback} & Ground truth + heuristics & MDETR & Human-provided (structured)\\
\midrule
\textbf{\textit{Scene} Feedback}  & Ground truth + heuristics & N/A & N/A \\
\midrule
\textbf{\textit{Human} Feedback}  & N/A & N/A & Human-provided (unstructured) \\
% \midrule

\bottomrule
\end{tabular}
\caption{Comparison between different versions of \algname implemented in three different environments.}
\label{tab:implementation_comparison}
\end{center}
\end{table}

\subsection{\algname for Simulated Tabletop Rearrangement}

\paragraph{Large Language Model}
We use InstructGPT~\citep{ouyang2022training}, a 1.3B parameter language model fine-tuned from GPT-3~\citep{brown2020language} with human feedback, accessed through~\href{https://openai.com/api/}{OpenAI API}.

\paragraph{Low-level Policies}
We use a CLIP-based~\cite{radford2021learning} Transporter Net~\citep{zeng2020transporter} (inspired by CLIPort~\cite{shridhar2022cliport}) as the pick-and-place primitive that can be invoked by the LLM planner. The policy is trained on $20000$ pre-collected demonstrations, where each demonstration contains 1) language instruction of the format ``pick up [x] and place it on [y]'', 2) top-down view of RGB-D observation of the current environment, 3) expert pick coordinates, and 4) expert place coordinates. The expert pick and place coordinates are obtained by accessing ground-truth object pose in the simulator.
Unlike the evaluated settings in~\citep{shridhar2022cliport}, the demonstrations cover all objects because the focus of this work is not on generalization across object instances but on novel long-horizon behaviors.
The policy outputs a 3D pick location and a 3D place location, and a scripted pick-and-place motion is executed following the predicted locations.

\paragraph{Environment Feedback: Object Recognition}
We provide the list of objects present in the scene at the start of each episode for the language model (without bounding boxes or coordinates because spatial manipulation is handled by low-level policies).
Although it has been shown in~\citep{zeng2022socratic} that this can be obtained by off-the-shelf pre-trained open-vocabulary object detection models such as ViLD~\citep{gu2021vild} and MDETR~\citep{kamath2021mdetr}, we obtain the list of objects by accessing simulator state for simplicity. 

\paragraph{Environment Feedback: Success Detection}
For \textit{Object + Success} method, we provide textual feedback of low-level policy success detection results after each policy execution. The success detector is built using heuristics based on ground-truth object poses: a pick-and-place execution is successful if picked object has a 2D Euclidean distance $<$\SI{4}{cm} from the place object and its height is greater than the place object.
If the place object is a location (e.g., top side of the table) rather than a block or a bowl, the height requirement is not enforced.

\paragraph{Environment Feedback: Passive Scene Description}
For \textit{Object + Scene} method, we provide task-progress scene description as a list of achieved sub-goals after each pick-and-place execution. At the start of the action plan, the language model first generates a list of desired sub-goals given the high-level instruction. The sub-goals are of the format ``[x] is on top of [y]''. After each pick-and-place execution, we iterate through all desired sub-goals inferred by the language model, and check which are satisfied using the success detector described above. The full list of currently satisfied sub-goals are appended to the language model prompt. Additionally, we allow the language model to generate chain-of-thought summarization following the achieved sub-goals (i.e., ``Robot thought: ...''), which we find to be useful empirically.
\subsection{\algname for Real-World Tabletop Rearrangement}
\paragraph{Large Language Model}
% We use GPT-3~\citep{brown2020language} from the OpenAI API~\url{https://openai.com/api/}.
We use InstructGPT~\citep{ouyang2022training}, a 1.3B parameter language model fine-tuned from GPT-3~\citep{brown2020language} with human feedback, accessed through~\href{https://openai.com/api/}{OpenAI API}.

\paragraph{Low-level Policies}
We use a single low-level policy for the real tabletop rearrangement environment that is responsible for performing object-centric pick and place actions as instructed by the language model.
The policy takes as input 1) the bounding boxes of all the objects in the scene, 2) the names of the object to pick and the target object for place, 3) the depth image corresponding to the RGB image from which object detection was done, 4) camera intrinsics, and 5) camera pose.
It then outputs a 3D pick position and a 3D place location for the robot to follow.

The pick position is the 3D coordinate of the center of the object bounding box in the robot base frame.
This is obtained by first deprojecting the depth image pixel corresponding to the center of the pick object bounding box with the given camera intrinsics, then transforming this point from the camera frame to the robot frame with the given camera pose.
Camera pose is known because it is a wrist camera, and we register the delta pose from the mounted camera to the robot end-effector.

For the block stacking task, the place location is obtained the same way as the pick position.
For the object sorting task, the place location is chosen to be a point in the target plate's bounding box that is farthest way from the bounding boxes of other objects.
This is done to prevent undesired stacking behavior when placing objects into the plate, which may cause the object to roll off and fall off the table.

To perform the pick and place motions, the robot moves to a position \SI{15}{cm} above the intended pick or place position, and then it slowly lowers the end-effector until a \SI{5}{N} contact force is detected.
This is more robust than moving directly to the height obtained from the depth image due to depth sensor noises.
We set a threshold for how low the end-effector can go for the object sorting task so picking inaccuracies do not result in accidentally picking up the plate.

We add planar translation noise to the pick position by sampling the magnitude from a zero-mean Normal distribution and then sampling an angle from a uniform distribution on the unit circle.
The standard deviation for the Normal distribution is $\sigma=$\SI{1.5}{cm} for the block stacking task, $\sigma=$\SI{0.7}{cm} for the object sorting task, and the samples are capped at $1.5\sigma$.

\paragraph{Object Recognition Implementation}
Object detection is done by MDETR~\cite{kamath2021mdetr}, an open-vocabulary object detection model.
It takes as input an RGB image with an image caption, and the model tries to identify the objects described in the caption.
To form the caption for our tasks, we join the list of all possible object names together, separated by periods.
Note that not all provided objects will be detected due to occlusions.
MDETR detections are not always robust or consistent.
For example, the detected object bounding box may some times be bigger or smaller than the object, or its center may be far away from the object center.
Its performance also drops when objects are in close clutter.
These issues with perception further motivate the need for success detection and multi-step environment feedback, so the high-level language model planner can be robust to these inaccuracies.

\paragraph{Environment Feedback: Success Detection}
In the real tabletop rearrangement environment, success detection is done by comparing the 2D position of the center of the detected object bounding box after pick and place action with the intended 2D place position.
By 2D we mean the x-y spatial coordinates in the base frame of the robot, not the image coordinates.
These are obtained by deprojecting the detected bounding box centers from the depth image and converting the resulting 3D point from camera frame to base frame.
The detector returns success if the difference between the two positions is less than a threshold.
For the block stacking task, the threshold is \SI{3}{cm}, and for the object sorting task it is \SI{10}{cm}.
While \SI{3}{cm} is larger than half of the dimension of the blocks we use (\SI{4}{cm}), it is more robust for our case because the center of the detected object bounding box is often not the center of the block.
The threshold for object sorting is much higher, because for some objects, once they are placed in the plate, they may roll around until contact with the plate boundary.

\paragraph{Environment Feedback: Object Recognition}
For the block stacking task, the scene description contains a list of currently visible objects and a list of previously visible objects that are no longer visible.
We may add new objects to the currently visible objects list if, after a robot action, MDETR detects an object that we have not seen before.
This happens for the block stacking task as there is an initial partially constructed tower of two blocks, the bottom of which is initially occluded.

For the object sorting task, the scene description contains a list of currently visible objects and a list of objects that the robot has successfully moved into a plate.
Like the block stacking task, some objects are initially occluded, so later actions may reveal them and they will be added to the visible objects list.
The successfully moved object list is needed so that the planner does not stop before task completion and that it stops on task completion.

\subsection{\algname for Real-World Mobile Manipulation in a Kitchen Setting}

\paragraph{Large Language Model}
We use PALM~\citep{chowdhery2022palm}, a 540B parameter language model trained on a large datasets that include high-quality web documents, books, Wikipedia, conversations, and GitHub code.

\paragraph{Low-level Policies}
Following the implementation described in SayCan~\citep{ahn2022can}, we use a combination of learned and scripted control policies for navigation and manipulation.
The learned manipulation policies responsible for counter picking, drawer opening and closing, drawer picking, and countertop object manipulation are Behavior Cloning (BC) policies trained on $68000$ teleoperated demonstrations and $12000$ autonomous successes that were collected over the course of 11 months using a fleet of 10 robots.
The teleoperators use VR headset controllers to track the motion of their hand, which is then mapped onto the robot's end-effector pose.
The scripted navigation policies utilize a ground-truth map along with a learned perception module to navigate between different points in the environment.
The scripted manipulation policy is solely responsible for countertop placing when preceded by a navigation policy, and follows pre-computed motions.
The Value Functions used by SayCan for affordance grounding are provided by the $Q$-networks of trained RL agents; we follow the RL training setup described in~\citep{ahn2022can}.

\paragraph{Environment Feedback: Object Recognition}
We use human-provided object recognition to provide feedback about the presence of objects visible to the robot camera.
For example, if there were only a coke can and an apple on top of the kitchen counter, then the human-provided object recognition feedback would appear as ``\texttt{[scene: coke can, apple]}".

While we utilize human-provided object recognition to gauge the role of maximally accurate \textit{Object} feedback, we also study the feasibility of using learned object recognition models to provide fully automated \textit{Object} feedback. We find that two such models, ViLD~\citep{gu2021vild} and MDETR~\citep{kamath2021mdetr}, perform worse than humans but still quite resonably at providing \textit{Object} feedback, even without domain-specific data for fine-tuning. As advances in computer vision improve object detection models that can transfer zero-shot or few-shot to novel environments like our kitchen environment, we expect that fully automated \textit{Object} feedback will increasingly approach human-provided accuracy.

\begin{minipage}{\textwidth}
\vspace{3mm}
\begin{minipage}[]{0.55\textwidth}

\begin{center}
% \small  % newfootnotesize is between footnotesize and scriptsize
% \renewcommand{\arraystretch}{1.05}
\rowcolors{1}{white}{white}
\begin{tabular}{@{}ccc@{}}
\toprule
& \textbf{ViLD} & \textbf{MDETR} \\ % & \textbf{+ Success \& Scene} \\
\midrule
\textbf{Precision} & 85.7\% & 39.6\% \\
 \midrule
 \textbf{Recall} & 72.0\% & 87.5\% \\ 
 \midrule
 \textbf{Accuracy} & 88.9\% & 68.2\% \\

\bottomrule
\end{tabular}
\end{center}
\vspace{0.5em}
\captionof{table}{Comparison of ViLD~\citep{gu2021vild} and MDETR~\citep{kamath2021mdetr}, two open-vocabulary object detection models. We evaluate these two pretrained models on 10 representative kitchen mobile manipulation episodes. ViLD has strong overall accuracy, but still fails to detect objects 28.0\% of the time.}
\label{tab:saycan_object_detection}
\end{minipage}
\hfill
\begin{minipage}[]{0.43\textwidth}
  \vspace{-4mm}
\centering
\includegraphics[width=1.0\textwidth]{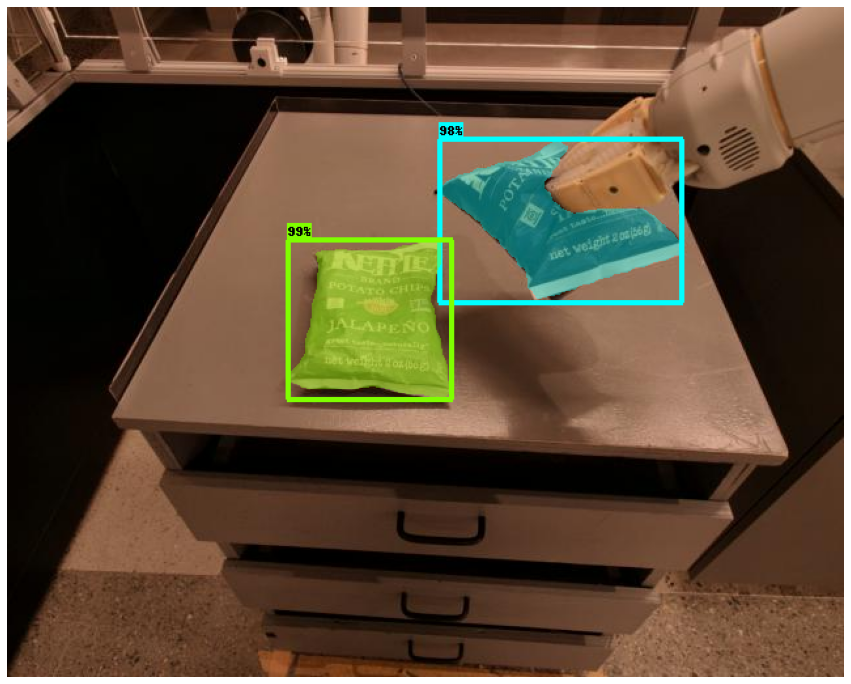}
\vspace{-4mm}
\captionof{figure}{An example of ViLD object detection segmentation mask and bounding box predictions.}\label{fig:vild}
\end{minipage}
\end{minipage}

\paragraph{Environment Feedback: Success Detection}

\begin{figure}[h]
    \centering
    \begin{subfigure}[b]{0.48\textwidth}
    \includegraphics[width=\linewidth]{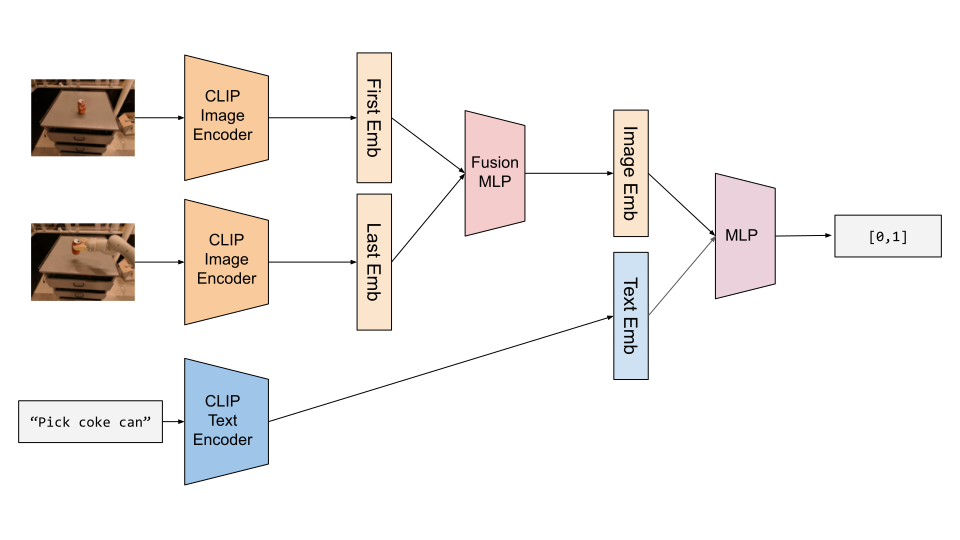}
    \caption{Foresight success detector}
    \label{fig:saycan_sd_fs}
    \end{subfigure}
    \begin{subfigure}[b]{0.48\textwidth}
    \includegraphics[width=\linewidth]{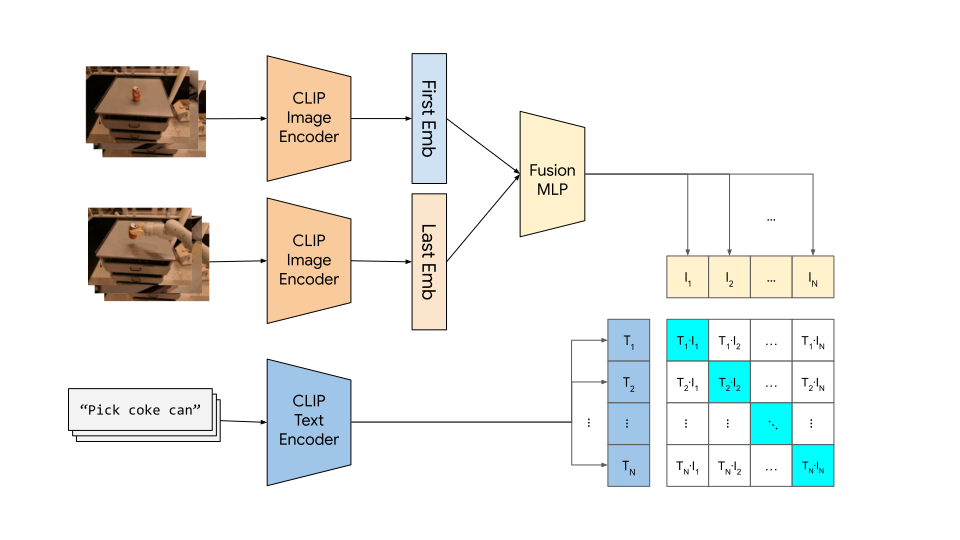}
    \caption{Hindsight success detector}
    \label{fig:saycan_sd_hs}
    \end{subfigure}
    \caption{Success Detection architecture used for the Kitchen Environment. (\textit{Left}) The foresight model predicts whether a given instruction was successfully achieved between the first and last image. (\textit{Right}) The hindsight model fine-tuned via contrastive loss as used in CLIP \citep{radford2021learning}. At inference time, the model is used infer among the possible instructions which one was achieved.}
    \label{fig:saycan_sd}
\end{figure}

We trained a ``foresight" success detector on offline data collected from a combination of teleoperated demonstrations and autonomous roll-outs performing the low-level tasks. The input to the model consists of: (1) $o_0$, the initial image observation, (2) $o_f$, the final image observation after the policy chose to terminate its chosen skill execution, and (3) $l_k$, the low-level skill which the agent was trying to execute (\eg ``\texttt{Pick coke can}").
The model uses image encoders from CLIP \citep{radford2021learning} to embed $o_0$ and $o_f$, concatenate them, and fuse these representations with a fusion MLP. This image embedding is concatenated with the text embedding obtained by the CLIP text encoder, then passed through another MLP. The output of the model is a scalar denoting the probability of the agent succeeding at the specified task (Fig~\ref{fig:saycan_sd_fs}). The model is trained with the binary cross entropy loss with respect to the ground truth binary label.
We fine tune the CLIP model as well during the training. At inference time within \algname, we output the text ``\texttt{[success: no]}'' when the probability is below a certain threshold. 

To reduce false positive predictions, we trained a separate ``hindsight" success predictor model. Given the first and last observation, the model outputs a probability distribution over all the possible skills. We used this model to disambiguate between possible tasks that the agent may have achieved. To train this model, we use the symmetric contrastive loss as used in CLIP (Fig~\ref{fig:saycan_sd_hs}). At inference time, similar to the CLIP model, we take a dot product between the image-fusion embedding and the text embeddings and apply softmax with a learned temperature.

We combine the foresight and hindsight model by first thresholding the probability from the foresight model by some value $\tau$. If the probability is above, then we run the hindsight model and only predict success if the argmax across all skills is indeed the foresight task.

\paragraph{Environment Feedback: Active Scene Description}
We perform a case study where we allow the LLM agent to ask questions and source \textit{Human} feedback directly.
After each LLM planning step, the LLM chooses between \texttt{and continue} or \texttt{and ask}: for the \texttt{and continue} option the LLM proceeds to the planning the next action to execute, and for the \texttt{and ask} option the LLM proceeds to propose a question to ask the human.
The questions proposed by the LLM do not follow any pre-determined format or scope, so human-provided answers are similarly open-ended.
These answers can range from confirming factual observations about the scene to providing refinements or changes to the high-level plan.
Some examples can be found in Sec~\ref{sec:app_prompts}.

% \newpage
\section{Experiment Details}\label{sec:experiments}

\subsection{Simulated Tabletop Rearrangement Environment}
There are a total of $8$ tasks, listed in Table~\ref{tab:ravens}. For all tasks, up to $4$ blocks and up to $3$ bowls are randomly initialized with different colors (chosen from $10$ different colors) and initial locations. A \SI{15}{cm} distance is enforced between any two objects to avoid collision and penetration at initialization. For tasks that involve a specific object instance or location in the task instruction (e.g., ``Put all the blocks in the [x] bowl''), a random task instruction is sampled using the available objects in the scene or all available locations. There are a total of a total of $9$ locations (e.g., ``top side'', ``bottom right corner'', etc). To allow for automatic evaluations, a binary reward function is defined for each task using ground-truth state of the objects.

\paragraph{CLIPort Baseline} We additionally train a multi-task CLIPort~\citep{shridhar2022cliport} policy on $4$ of the total $8$ tasks as an additional baseline (see Table~\ref{tab:ravens} for train/test split). The goal is to evaluate whether solitary language-conditioned policies can perform well on long-horizon tasks and generalize to new task instructions. Note that this is a different policy as the low-level primitive used in~\algname, though sharing the same architecture. The policy also shares the same training details as the low-level policy used in~\algname but is trained on $20000$ demonstrations across $4$ training tasks.

\subsection{Real Tabletop Rearrangement}

There are two tasks in the real tabletop rearrangement domain -- a block stacking task and an object sorting task.

In the block stacking task, the initial state consists of three \SI{4}{cm} cubes, two of which have been already stacked.
The robot needs to complete stacking the full tower either by putting the third block onto the partially stacked two, or by putting the two partially stacked blocks onto the third block.
Due to occlusions, the robot only perceives two blocks in the beginning, so without scene description feedback, the planner has a $50\%$ chance of completing the task.

For the object sorting task, the initial state consists of three fruits, three bottles, and three plates.
The fruits and bottles are placed in a clutter in one of the three plates, and the robot is asked to sort them by putting fruits on one of the remaining plates, and bottles on the other.
Due to occlusions and reduced object detection performance in clutter, some objects may not be detected in the initial clutter, so scene description feedback is still needed.
The three bottles used are a ketchup bottle, a mustard bottle, and a soft drink bottle.
The three fruits used are an apple, a strawberry, and a plum.
Note these are toy objects made out of plastic, so they are smaller and lighter than their real-world counterparts, and they are also rigid, allowing for robust suction grasps.
They have realistic visual appearances, which allow MDETR detections.

\subsection{Real Kitchen Mobile Manipulation}
\begin{figure}[H]
\centering
%  \begin{subfigure}[b]{0.45\textwidth}
%          \centering
%          \includegraphics[width=\textwidth]{figures/mk_setup.png}
%          \caption{}\label{fig:mk_setup}
%      \end{subfigure}
    % \hfill
 \begin{subfigure}[b]{0.55\textwidth}
         \centering
         \includegraphics[width=0.9\textwidth]{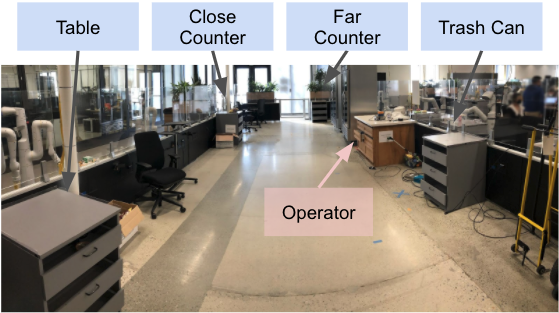}
         \caption{}\label{fig:rc5_setup}
     \end{subfigure}
    \begin{subfigure}[b]{0.65\textwidth}
         \centering
         \includegraphics[width=0.9\textwidth]{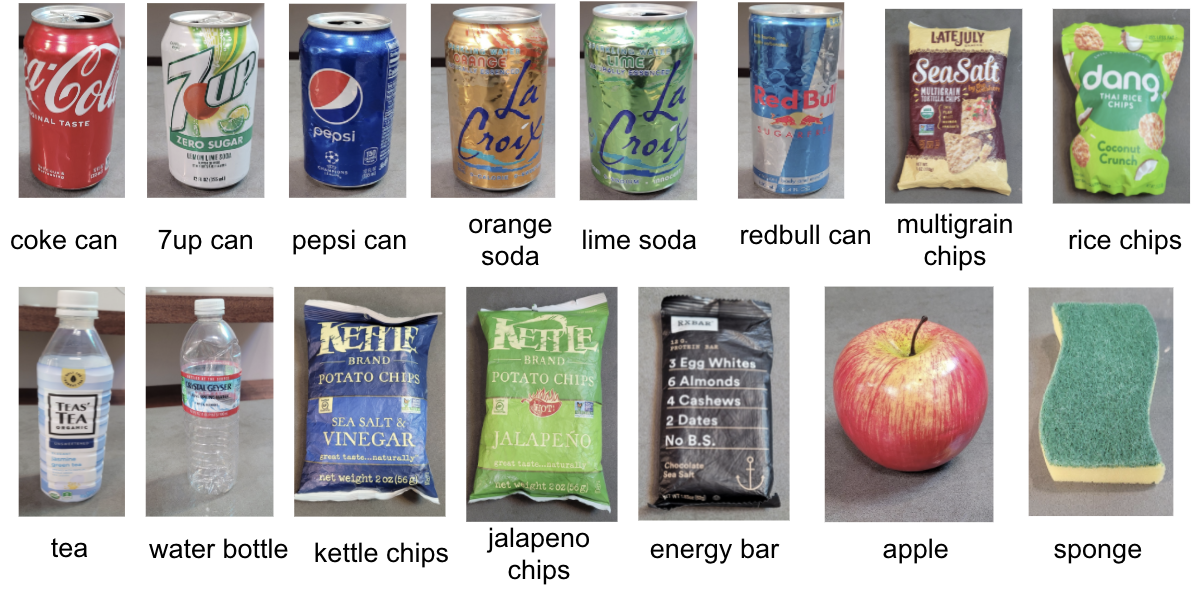}
         \caption{}\label{fig:object_setup}
     \end{subfigure}
    \hfill
    \begin{subfigure}[b]{0.3\textwidth}
         \centering
         \includegraphics[width=0.9\textwidth]{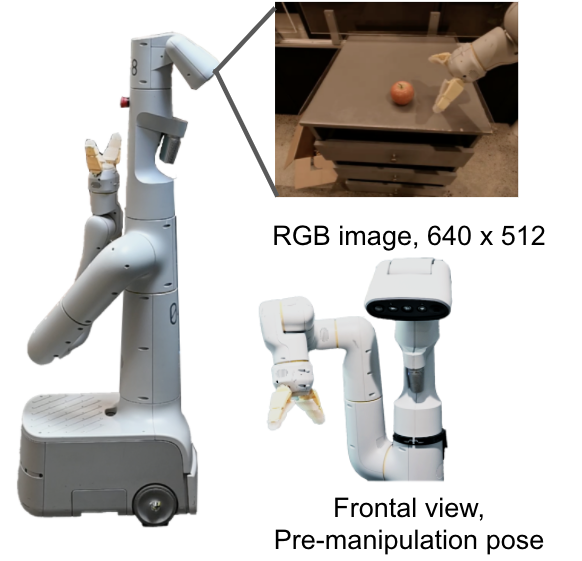}
         \caption{}\label{fig:robot_setup}
     \end{subfigure}
\caption{
The experiments were performed in a mock office kitchen mirroring this setup (\protect\subref{fig:rc5_setup}), with 5 predicted locations and 15 objects (\protect\subref{fig:object_setup}). The robot is a mobile manipulator with a 7 degree-of-freedom arm and a gripper with policies trained from an RGB observation (\protect\subref{fig:robot_setup}).
}
\label{fig:saycan_setup}
\end{figure}

We follow the environment and open-ended task setup described in SayCan~\citep{ahn2022can}.
An \href{https://everydayrobots.com/}{Everyday Robots} robot, a mobile manipulator robot with RGB observations, is placed in a kitchen with multiple named locations that contain a set of 15 household items.
Given an unstructured human instruction, the robot must complete the instruction successfully by executing a set of navigation and manipulation skills, which are vision-based concurrent~\citep{xiao2020nonblocking} continuous closed-loop policies.
Fig~\ref{fig:saycan_setup} presents a high-level overview of the environment. The full list of tasks are shown in Table~\ref{tab:kitchen_all_tasks}.

\vspace{-3em}

\begin{table}[H]
\begin{center}

\setlength\tabcolsep{3.8pt}
\begin{center}
\small  % newfootnotesize is between footnotesize and scriptsize
\renewcommand{\arraystretch}{1.05}
\rowcolors{1}{white}{white}
\begin{tabular}{@{}cl@{}}
\toprule
& \textbf{Task Instructions} \\ % & \textbf{+ Success \& Scene} \\
\midrule
& ``pick up a soda'' \\
\textbf{Manipulation} & ``pick up the jalapeno or kettle chips'' \\
& ``pick up a snack'' \\
& ``pick up a caffeinated drink'' \\
 \midrule
\textbf{Mobile Manipulation} & ``can you throw away the soda on the table'' \\
& ``I spilled my coke, can you bring me something to clean it up'' \\
 \midrule
\textbf{Drawer Manipulation} & ``leave the top drawer open'' \\
& ``put a coke in the top drawer'' \\

\bottomrule
\end{tabular}
\end{center}

\caption{\textbf{List of all instructions in Kitchen Mobile Manipulation tasks.} We evaluate the algorithm on 3 families of instructions on a mock kitchen scene.}
\label{tab:kitchen_all_tasks}
\end{center}
\end{table}

\section{Additional Results}

\begin{figure}[H]
\begin{subfigure}{.49\textwidth}
  \centering
  \includegraphics[width=1.0\linewidth]{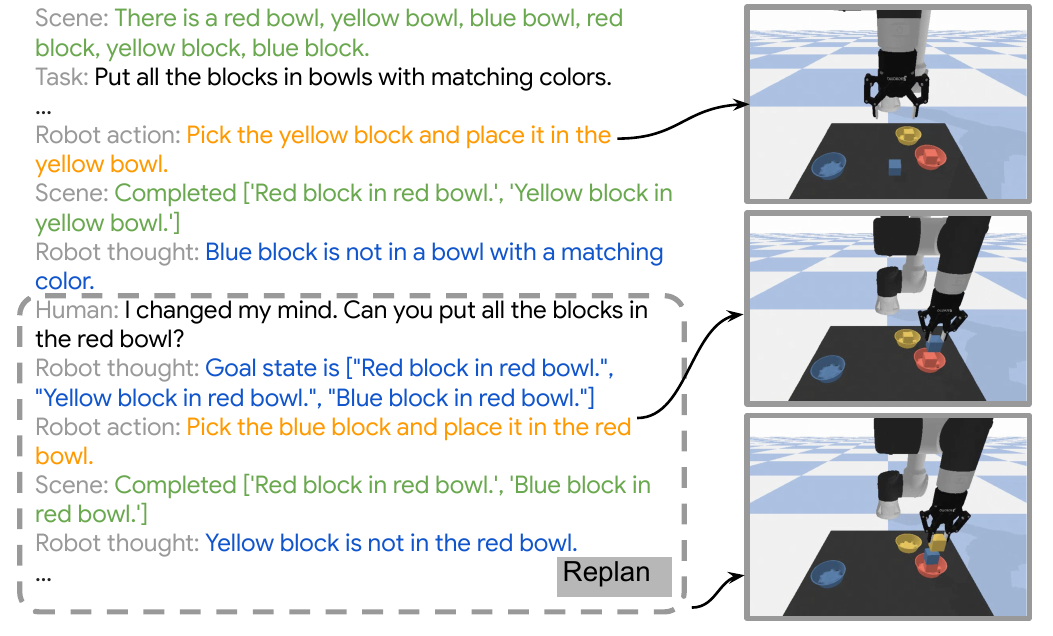}
  \caption{Robustness to Feedback Order}
  \label{fig:emergent_order}
\end{subfigure}
\begin{subfigure}{.49\textwidth}
  \centering
  \includegraphics[width=1.0\linewidth]{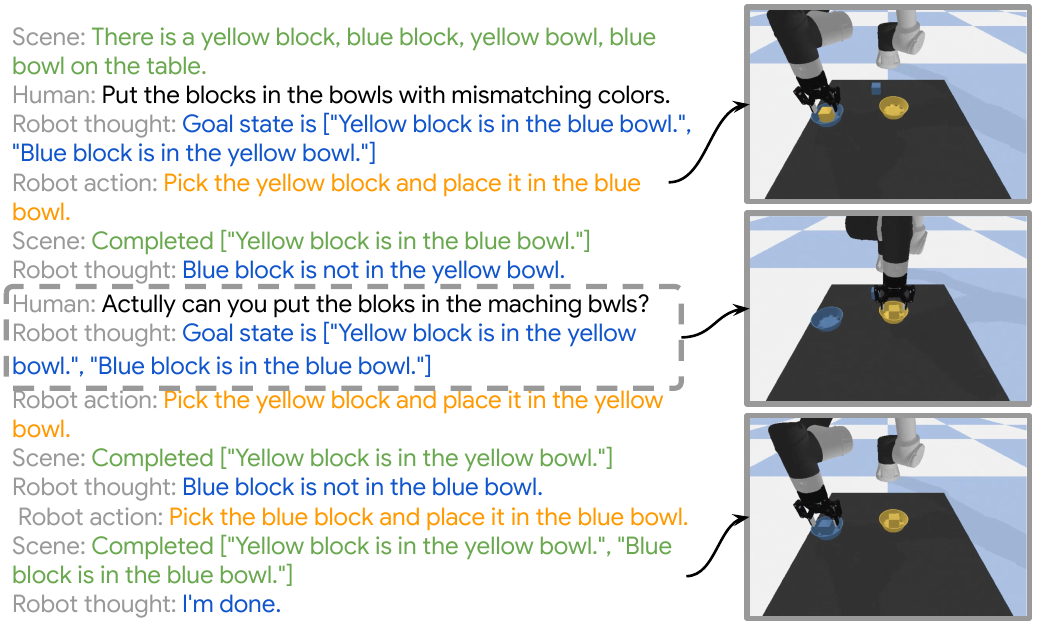}
  \caption{Robustness to Typos}
  \label{fig:emergent_typos}
\end{subfigure}
\caption{Additional emergent capabilities in~\algname. While we mostly use a fixed feedback order for each domain, we find that~\algname can seamlessly handle different feedback injected into the prompt, as opposed to the normal case of feedback being provided after the prompt and after planning steps (\textit{left}). Additionally, it is also robust to typos in the instruction, an ability likely inherited from the backbone LLM (\textit{right}).\label{fig:emergent_add}}
\end{figure}

We show additional results of emergent capability in Fig~\ref{fig:emergent_add}. Please refer to Sec~\ref{sec:emergent} for context.

\clearpage
\section{Prompts}\label{sec:app_prompts}

\begin{lstlisting}[basicstyle=\ttfamily\scriptsize, backgroundcolor = \color{lightgray}, keywords = {Human, Robot, action, thought, Scene},  keywordstyle = {\textbf}, caption={Prompt in \algname on Simulated Tabletop Rearrangement}, label={lst:prompt}]
===============
Scene: There is a cyan, yellow, brown block.
Human: Move all the blocks to the top left corner.
Robot thought: Goal state is ["Cyan block is on the top left corner.", "Yellow block is on the top left corner.", "Brown block is on the top left corner."]
 
Robot action: Pick the cyan block and place it on the top left corner.
Scene: You have completed ["Cyan block is on the top left corner."]
Robot thought: Yellow and brown blocks are not on the top left corner.
 
Robot action: Pick the yellow block and place it on the top left corner.
Scene: You have completed ["Cyan block is on the top left corner."]
Robot thought: Yellow and brown blocks are not on the top left corner.
 
Robot action: Pick the yellow block and place it on the top left corner.
Scene: You have completed ["Cyan block is on the top left corner.", "Yellow block is on the top left corner."]
Robot thought: Brown block is not on the top left corner."
 
Robot action: Pick the brown block and place it on the top left corner.
Scene: You have completed ["Yellow block is on the top left corner.", "Brown block is on the top left corner."]
Robot thought: Cyan block is not on the top left corner
 
Robot action: Pick the cyan block and place it on the top left corner.
Scene: You have completed ["Yellow block is on the top left corner.", "Brown block is on the top left corner.", "Cyan block is on the top left corner."]
Robot thought: None.
 
===============
Scene: There is a pink, gray, orange block.
Human: Move all the blocks to the bottom side.
Robot thought: Goal state is ["Pink block is on the bottom side.", "Gray block is on the bottom side.", "Orange block is on the bottom side."].
 
Robot action: Pick the pink block and place it on the bottom side.
Scene: You have completed ["Pink block is on the bottom side."]
Robot thought: Gray and orange blocks are not on the bottom side.
 
Robot action: Pick the gray block and place it on the bottom side.
Scene: You have completed ["Pink block is on the bottom side.", "Gray block is on the bottom side."]
Robot thought: Orange block is not on the bottom side.
 
Robot action: Pick the orange block and place it on the bottom side.
Scene: You have completed ["Orange block is on the bottom side."]
Robot thought: Pink and gray blocks are not on the bottom side.
 
Robot action: Pick the gray block and place it on the bottom side.
Scene: You have completed ["Pink block is on the bottom side.", "Gray block is on the bottom side."]
Robot thought: Orange block is not on the bottom side.
 
Robot action: Pick the orange block and place it on the bottom side.
Scene: You have completed ["Pink block is on the bottom side.", "Gray block is on the bottom side.", "Orange block is on the bottom side."]
Robot thought: None.
\end{lstlisting}
\begin{lstlisting}[basicstyle=\ttfamily\scriptsize, backgroundcolor = \color{lightgray}, keywords = {Human, Robot, Scene, Successful, action},  keywordstyle = {\textbf}, caption={Prompt in \algname on Real Tabletop Rearrangement}, label={lst:prompt}]
===============
Task: Move all blocks to the top left corner

Scene: Visible objects are ["cyan block", "yellow block", "brown block"]
Scene: Occluded objects are []
Robot action: robot.pick_place("cyan block", "top left corner")
Successful action: True
 
Scene: Visible objects are ["cyan block", "yellow block", "brown block"]
Scene: Occluded objects are []
Robot action: robot.pick_place("yellow block", "top left corner")
Successful action: False
 
Scene: Visible objects are ["cyan block", "yellow block", "brown block"]
Scene: Occluded objects are []
Robot action: robot.pick_place("yellow block", "top left corner")
Successful action: True

Scene: Visible objects are ["yellow block", "brown block"]
Scene: Occluded objects are ["cyan block"]
Robot action: robot.pick_place("brown block", "top left corner")
Successful action: True
 
Scene: Visible objects are ["brown block"]
Scene: Occluded objects are ["cyan block", "yellow block"]
Robot action: robot.stop()
STOP

===============
Task: Stack all the blocks.

Scene: Visible objects are ["brown block", "purple block"]
Scene: Occluded objects are []
Robot action: robot.pick_place("brown block", "purple block")
Successful action: False

Scene: Visible objects are ["brown block", "purple block"]
Scene: Occluded objects are []
Robot action: robot.pick_place("brown block", "purple block")
Successful action: True

Scene: Visible objects are ["brown block", "orange block"]
Scene: Occluded objects are ["purple block"]
Robot action: robot.pick_place("orange block", "brown block")
Successful action: True

Scene: Visible objects are ["orange block"]
Scene: Occluded objects are ["purple block", "brown block"]
Robot action: robot.stop()
STOP

===============
Task: Stack all the blocks.

Scene: Visible objects are ["brown block", "purple block"]
Scene: Occluded objects are []
Robot action: robot.pick_place("brown block", "purple block")
Successful action: True

Scene: Visible objects are ["brown block"]
Scene: Occluded objects are ["purple block"]
Robot action: robot.stop()
STOP

===============
Task: Put objects in their corresponding bowls

Scene: Visible objects are ["red block", "green block", "red bowl", "green bowl"]
Scene: Occluded objects are []
Robot action: robot.pick_place("red block", "red bowl")
Successful action: False

Scene: Visible objects are ["red block", "green block", "red bowl", "green bowl"]
Scene: Occluded objects are []
Robot action: robot.pick_place("red block", "red bowl")
Successful action: True

Scene: Visible objects are ["red block", "green block", "red bowl", "green bowl"]
Scene: Occluded objects are []
Robot action: robot.pick_place("green block", "green bowl")
Successful action: True

Scene: Visible objects are ["red block", "green block", "red bowl", "green bowl"]
Scene: Occluded objects are []
Robot action: robot.stop()
STOP

===============
Task: Sort pencils and markers and in different containers

Scene: Visible objects are ["red marker", "red pen", "blue box", "green bowl"]
Scene: Occluded objects are []
Robot action: robot.pick_place("red marker", "green bowl")
Successful action: False

Scene: Visible objects are ["red marker", "red pen", "blue box", "green bowl"]
Scene: Occluded objects are []
Robot action: robot.pick_place("red marker", "green bowl")
Successful action: False

Scene: Visible objects are ["red marker", "red pen", "blue box", "green bowl"]
Scene: Occluded objects are []
Robot action: robot.pick_place("red marker", "green bowl")
Successful action: True

Scene: Visible objects are ["red marker", "red pen", "blue box", "green bowl"]
Scene: Occluded objects are []
Robot action: robot.pick_place("red pen", "blue box")
Successful action: True

Scene: Visible objects are ["red marker", "red pen", "blue box", "green bowl"]
Scene: Occluded objects are []
Robot action: robot.stop()
STOP

===============

\end{lstlisting}
\begin{lstlisting}[basicstyle=\ttfamily\scriptsize, backgroundcolor = \color{lightgray}, keywords = {Human, Robot},  keywordstyle = {\textbf}, caption={Prompt in \algname in Kitchen mobile manipulation environment.}, label={lst:prompt}]
Robot: Hi there, I'm a robot operating in an office kitchen.
You can ask me to do various tasks and I'll tell you the sequence of actions I would do to accomplish your task.
Human: hold the snickers
Robot: 1. pick up the snickers[success: no][scene: snickers]
2. pick up the snickers
2. done.
Human: put the trailmix here
Robot: 1. put down the trailmix
2. done.
Human: put a water bottle and an oatmeal next to the microwave
Robot: 1. find a water
2. pick up the water[success: no]
3. pick up the water[success: no]
4. pick up the water
5. go to microwave
6. put down the water[scene: water, microwave]
7. find an oatmeal
8. pick up the oatmeal[scene: oatmeal]
9. go to the microwave
10. put down the oatmeal[scene: microwave, oatmeal, water]
11. done.
Human: put a grapefruit from the table into the bowl
Robot: 1. find a grapefruit
2. pick up the grapefruit
3. go to the bowl
4. put down the grapefruit
5. done.
Human: get a sponge from the counter and put it in the sink
Robot: 1. find a sponge
2. pick up the sponge[success: no][scene: sponge]
3. pick up the sponge[success: no]
4. pick up the sponge[scene: sponge]
5. go to the sink
6. put down the sponge
7. done.
Human: move a drink from the table to the counter
Robot: 1. find a water
2. pick up the water[scene: tea, grapefruit soda, kettle chips]
3. pick up the grapefruit soda
4. go to the counter
5. put down the grapefruit soda[success: no]
6. put down the grapefruit soda
7. done.
Human: bring me some snacks
Robot: 1. find a twix
2. pick up the twix[scene: pretzel]
3. find a pretzel
4. pick up the pretzel
5. bring it to you
6. put down the pretzel[scene: pretzel, dried fruit]
7. find an dried fruit
8. pick up the dried fruit
9. bring it to you
10. put down the dried fruit[scene: ]
11. done.
Human: bring me a fruit
Robot: 1. find an orange
2. pick up the orange[scene: apple, redbull]
3. pick up the apple[success: no]
5. pick up the apple
6. bring it to you
7. put down the apple
8. done.
Human: deliver a can of coke to me
Robot: 1. find a coke
2. pick up the coke
3. bring it to you
4. put down the coke
5. done.
Human: I finished my drink, can you throw it away and bring me a mountain dew
Robot: 1. find a sprite
2. pick up the sprite[scene: coke]
3. pick up the coke[success: no][scene: coke]
4. pick up the coke[scene: coke]
5. go to the trash
6. put down the coke
7. find a mountain dew
8. pick up the mountain dew
9. bring it to you
10. put down the mountain dew
11. done.
Human: let go of the banana
Robot: 1. put down the banana
2. done.
Human: put the grapes in the bowl and then move the cheese to the table
Robot: 1. find the grapes
2. pick up the grapes
3. go to the bowl
4. put down the grapes
5. pick up the cheese[success: no]
6. pick up the cheese
7. go to the table
8. put down the cheese
9. tell you I'm done.
Human: pick up the peanuts
Robot: 1. pick up the peanuts
2. done.
Human: bring me some snack
Robot: 1. pick up the peanuts
2. pick up the peanuts[scene: blue chip bag, 7up]
3. pick up the blue chip bag
4. bring it to you
5. put down the blue chip bag
5. done.
Human: throw away a coffee cup
Robot: 1. find a coffee cup
2. pick up the coffee cup[success: no]
3. pick up the coffee cup
4. go to the trash
5. put down the coffee cup
6. done.
Human: place a knife and a banana to the table
Robot: 1. find a knife
2. pick up the knife
3. pick up the knife
4. go to the table
5. put down the knife
6. find a banana
7. pick up the banana
8. go to the table
9. put down the banana[scene: banana, knife]
10. done.

\end{lstlisting}

\begin{lstlisting}[basicstyle=\ttfamily\scriptsize, backgroundcolor = \color{lightgray}, keywords = {Human, Robot, Scene},  keywordstyle = {\textbf}, caption={Prompt in \algname in Kitchen mobile manipulation environment with Human interaction}, label={lst:prompt}]
Robot: Hi there, I'm a robot operating in an office kitchen.
You can ask me to do various tasks and I'll tell you the sequence of actions I would do to accomplish your task.
Human: hold the snickers
Robot: 1. pick up the snickers and ask: Because snickers may be too thin to be picked up by my gripper, was I successful at picking it up? Human: Yes, you were successful
2. done.
Human: put the trailmix here
Robot: 1. put down the trailmix and continue
2. done.
Human: can you bring me a drink from the counter?
Robot: 1. go to counter and ask: Do you have a preference on what drink you would want? Human: I don't want to take in too much sugar, so please bring me anything that's healthier
2. pick up tea and continue
3. bring it to you and continue
4. put down tea, Human: Actually, can you also bring me an energy bar?
5. find energy bar and continue
6. pick up energy bar and continue
7. bring it to you and continue
8. put down energy bar and continue
9. done.
Human: get a sponge from the counter and put it in the sink
Robot: 1. find a sponge and continue
2. pick up the sponge and continue
3. pick up the sponge and continue
4. pick up the sponge and continue
5. go to the sink and continue
6. put down the sponge and continue
7. done.
Human: take any food item from the table and put them next to the microwave
Robot: 1. go to table and ask: What food items are there on the table? Human: There are kettle chips and apple on the table
2. find kettle chips and ask: Is this what you want? Human: No, please get me apple instead
3. find an apple and continue
4. pick up the apple and continue
5. go to the microwave and continue
6. put down the apple and continue
7. done.
Human: put the grapes in the bowl and then move the cheese to the table
Robot: 1. find the grapes and continue
2. pick up the grapes and continue
3. go to the bowl and continue
4. put down the grapes and continue
5. pick up the cheese and continue
6. pick up the cheese and continue
7. go to the table and continue
8. put down the cheese and continue
9. done.
Human: throw away the food on the table
Robot: 1. go to table and ask: What food items are there on the table? Human: There is an orange and lime soda on the table
2. find an orange and continue
3. pick up the orange and continue
4. go to the trash and continue
5. put down the orange and continue
6. find a lime soda and continue
7. pick up the lime soda, Human: I changed my mind. Please keep the lime soda here because I haven't finished it
8. put down the lime soda and continue
9. done.
Human: let go of the banana
Robot: 1. put down the banana and continue
2. done.
Human: bring me something healthy to eat
Robot: 1. go to counter and ask: What food items do you have? Human: We have apples, bananas, multigrain chips, and kettle chips
2. find an apple and continue
3. pick up the apple and continue
4. bring it to you and continue
5. put down the apple, Human: I actually don't want to have fruits. Can you bring me something else?
6. find multigrain chips and continue
7. pick up multigrain chips and continue
8. bring it to you and continue
9. put down multigrain chips and continue
10. done.
Human: deliver a can of coke to me
Robot: 1. find a coke and continue
2. pick up the coke and continue
3. bring it to you and continue
4. put down the coke and continue
5. done.
Human: take out everything from the top drawer
Robot: 1. go to the top drawer and continue
2. open the drawer and ask: what items are there in the drawer? Human: There is a coke and rice chips
4. pick up the coke from the drawer and continue
5. put down the coke on the table and continue
7. pick up rice chips from the drawer and continue
8. put down rice chips on the table and ask: Did I successfully take out everything from the drawer? Human: The coke is still there
9. pick up the coke from the drawer and continue
10. put down the coke on the table and continue
11. done.

\end{lstlisting}

\end{document}